\pdfoutput=1

\documentclass[11pt]{article}

\usepackage[preprint]{acl}

\usepackage{times}
\usepackage{latexsym}

\usepackage[T1]{fontenc}

\usepackage[utf8]{inputenc}

\usepackage{microtype}

\usepackage{inconsolata}

\usepackage{graphicx}
\usepackage{subcaption}
\usepackage{url}
\usepackage[htt]{hyphenat}
\usepackage{siunitx}
\usepackage{enumitem}
\usepackage{array}
\usepackage{tabularx}
\usepackage{nameref}
\usepackage{booktabs}
\usepackage{multirow}
\usepackage{rotating}
\definecolor{okabe_red}{HTML}{D55E00}
\definecolor{okabe_blue}{HTML}{0072B2}
\definecolor{okabe_green}{HTML}{009E73}
\definecolor{okabe_orange}{HTML}{E69F00}
\definecolor{okabe_purple}{HTML}{CC79A7}

\definecolor{dodgerblue}{HTML}{1E90FF}
\definecolor{darkred}{HTML}{8B0000}
\definecolor{darkgreen}{HTML}{006400}
\definecolor{purple}{HTML}{800080}
\definecolor{darkorange}{HTML}{FF8C00}
\definecolor{darkslategray}{HTML}{2F4F4F}
\definecolor{firebrick}{HTML}{B22222}
\definecolor{darkblue}{HTML}{00008B}
\definecolor{darkmagenta}{HTML}{8B008B}
\definecolor{forestgreen}{HTML}{228B22}
\definecolor{brown}{HTML}{A52A2A}
\definecolor{cadetblue}{HTML}{5F9EA0}
\definecolor{darkorchid}{HTML}{9932CC}
\definecolor{crimson}{HTML}{DC143C}
\definecolor{darkslateblue}{HTML}{483D8B}
\definecolor{deeppink}{HTML}{FF1493}

\usepackage[scaled=0.85]{helvet}
\newcommand{\narrowfont}[1]{{\fontfamily{phv}\selectfont #1}}

\title{The Prompt Makes the Person(a): A Systematic Evaluation of Sociodemographic Persona Prompting for Large Language Models}

\author{
    Marlene Lutz\textsuperscript{1}, Indira Sen\textsuperscript{1}, Georg Ahnert\textsuperscript{1}, Elisa Rogers\textsuperscript{1}, Markus Strohmaier\textsuperscript{1,2,3}\\
    \textsuperscript{1}University of Mannheim, 
    \textsuperscript{2}GESIS - Leibniz Institute for the Social Sciences,\\
    \textsuperscript{3}Complexity Science Hub Vienna\\
    \texttt{\{marlene.lutz}, \texttt{indira.sen}, \texttt{georg.ahnert}, \texttt{markus.strohmaier\}@uni-mannheim.de}\\
    \texttt{elisa.marie-rogers@students.uni-mannheim.de}
}

\begin{document}
\maketitle
\begin{abstract}
Persona prompting is increasingly used in large language models (LLMs) to simulate views of various sociodemographic groups. However, how a persona prompt is formulated can significantly affect outcomes, raising concerns about the fidelity of such simulations. 
Using five open-source LLMs, we systematically examine how different persona prompt strategies, specifically \textit{role adoption} formats and \textit{demographic priming} strategies, influence LLM simulations across 15 intersectional demographic groups in both open- and closed-ended tasks. 
Our findings show that LLMs struggle to simulate marginalized groups
but that the choice of demographic priming and role adoption strategy significantly impacts their portrayal.
Specifically, we find that prompting in an \textit{interview}-style format and \textit{name}-based priming can help reduce stereotyping and improve alignment. 
Surprisingly, smaller models like OLMo-2-7B outperform larger ones such as Llama-3.3-70B.
Our findings offer actionable guidance for designing sociodemographic persona prompts in LLM-based simulation studies.

\end{abstract}

\section{Introduction}

\begin{figure}[t!]
    \centering
    \includegraphics[width=\linewidth]{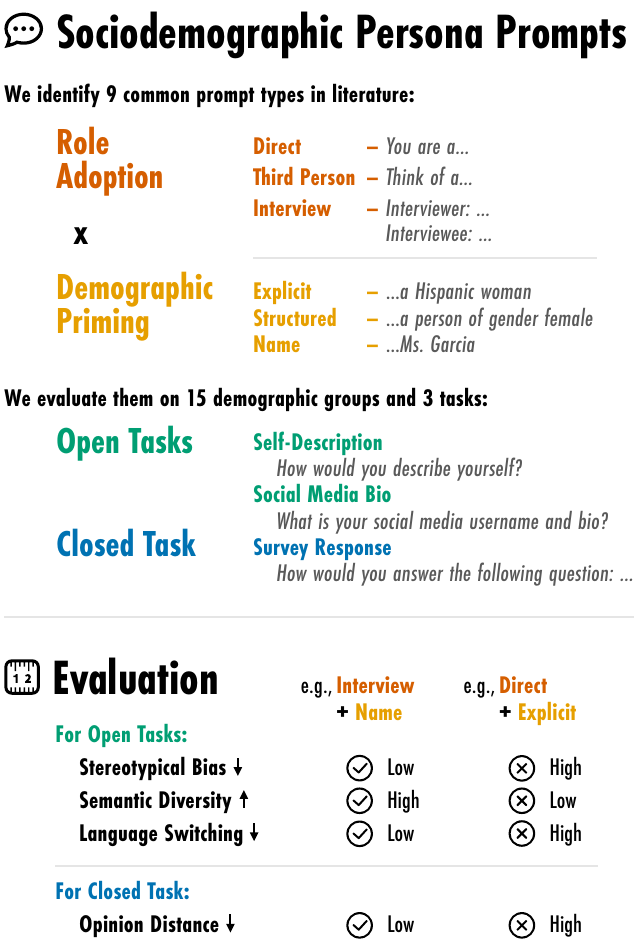}
    \caption{\textbf{Evaluation Framework for Sociodemographic Persona Prompting.} We construct sociodemographic persona prompts using combinations of three different \textit{role adoption} formats and three strategies for \textit{demographic priming}. We populate these prompts in conjunction with various sociodemographic groups and systematically evaluate them across both open- and closed-ended tasks using a broad set of bias and alignment measures.}
    \label{fig:figure1}
    \vspace{-1em}
\end{figure}

An increasing number of studies employ persona prompts to replace or supplement human input in social surveys, predict voting behavior, or perform subjective annotation tasks~\cite{santurkar2023whose,argyle2023out,beck2024sensitivity}. Persona prompting intends to condition a models' output to reflect the characteristics of specific personas, enabling researchers to simulate opinions, values, and attitudes more effectively. However, LLMs are known to be susceptible to seemingly minor prompt variations~\citep{zhou2024batch}, and prior work shows that outcomes can vary significantly depending on prompt formulation~\citep{beck2024sensitivity}.  
In line with~\citet{kovavc2023large}, we hypothesize that subtle contextual differences in persona prompts can activate different perspectives in LLMs, which, in turn, may influence their downstream behavior. This may occur because small changes in phrasing lead the model to draw on different aspects of its learned knowledge, depending on which associations are most strongly triggered by the input. As a concrete example, a model may simulate a group better if the demographic attributes are implicitly induced into the LLM, e.g., based on names containing demographic cues, rather than through explicit demographic descriptors which can provoke stereotype effects~\citep{spencer2016stereotype}.

However, research on sociodemographic persona prompting is in a nascent and exploratory phase where it remains unclear how different persona prompt strategies affect the representation of the subpopulations being modeled. The lack of clear guidelines 
has led to considerable variation in the articulation of demographic dimensions (\narrowfont{"Asian woman"}~\cite{cheng2023marked} vs. \narrowfont{"Ms. Huang"}~\cite{aher2023using}), and in how the persona’s point of view is constructed (\narrowfont{"A person can be described as follows: Race: \textit{X}"}~\cite{do2025aligning} vs. \narrowfont{"You are a person of race \textit{X}"}~\cite{beck2024sensitivity}).

To inform the design of our experiments, we conduct a survey of prompt strategies commonly used in LLM-based simulations studies. 
We identify various \textit{role adoption} formats, i.e., ways of prompting the model to take on a role 
--- through direct role assignment (\narrowfont{"You are a Black woman"}), using the third person singular (\narrowfont{"Think of a Black woman"}), or through a structured format where identity elements are introduced in an interview-like format (\narrowfont{"Interviewer: What is your race? Interviewee: My race is Black"}). 
Additionally, we examine the impact of \textit{demographic priming}, i.e., how demographic attributes are conveyed to the model --- implicitly via names (\narrowfont{"Ms. Gonzalez"}), explicitly via demographic terms (\narrowfont{"Hispanic woman"}) or explicitly via demographic categories and terms (\narrowfont{"a person of race Hispanic, and gender female"}). 
Using this framework, we evaluate how well LLMs represent demographic groups in both open-ended and closed-ended tasks --- 
 demographic biases in self-descriptions and bios and opinion alignment on survey questions~\cite{santurkar2023whose}, respectively (cf. Figure \ref{fig:figure1}). 

\textbf{Research questions.} 
We specifically study:

\begin{enumerate}[label=\textbf{(\roman*)}, itemsep=0.1em, topsep=0.2em]
    \item \textbf{Demographic Representativeness:} Is there a difference in LLMs' ability to simulate different demographic groups?
    \item \textbf{Strategies for Persona Prompting:} How do different persona prompt types impact LLMs' simulation abilities?
\end{enumerate}

\textbf{Results.} We find that LLMs do not simulate all sociodemographic groups equally well. In particular, prompts that invoke nonbinary, Hispanic, and Middle Eastern personas tend to elicit more stereotypical responses than for other personas. However, when systematically comparing different persona prompting strategies, we observe that \textit{interview} style and \textit{name}-based prompting result in less stereotypical associations, better opinion alignment, and reduced disparities between groups. Interestingly, larger models such as Llama-3.3-70B are less effective at simulating demographic groups, whereas the best simulation performance is achieved by OLMo-2-7B. 

\section{Sociodemographic Persona Prompting} \label{sec:prompt_format}
Sociodemographic persona prompting refers to a prompting technique that is used to steer the behavior of an LLM to align with that of a specified sociodemographic group or person~\citep{beck2024sensitivity}. 
We devise a framework to systematically analyze how different persona prompt types affect the simulation of sociodemographic groups in LLMs. 

We begin by identifying common prompt types from prior studies, which we then apply to simulate a variety of sociodemographic personas in both open- and closed-ended tasks.
As our foundation, we draw on the dataset curated by~\citet{sen_2025}, which compiles studies examining demographic representativeness in LLMs and includes annotations indicating whether a sociodemographic prompting approach was used. From this dataset, we sample 47 papers and extract their original prompts. 
Our review reveals recurring patterns in persona prompt types across studies, which we distill into two primary axes of variation (see Appendix \ref{app:lit_review} for further details on the review):
\begin{itemize}[itemsep=0.3em, topsep=0.3em, parsep=0em, partopsep=0em]
    \item \textit{Role Adoption:} the format in which the perspective of the persona is induced 
    \item \textit{Demographic Priming:} the descriptors used to signal the demographics of the persona
\end{itemize}
Based on these dimensions, we construct sociodemographic prompts and systematically examine their effect on LLM behavior across several tasks.

\subsection{Role Adoption Formats}\label{sec:role_adoption} 
We hypothesize that the format in which the model is instructed to adopt the role of a sociodemographic persona, along with the implied perspective (e.g., responding \textit{as} a persona versus \textit{about} a persona), can trigger different narratives that influence the portrayal of a persona.
Specifically, we differentiate the following role adoption formats:

\paragraph{Direct.}
The model is directly instructed to adopt a role and to respond as the assigned persona (e.g., used by \citet{gupta2024personabias,qu2024performance,hu2024quantifying}). An example would be: \narrowfont{"You are a Hispanic woman."}

\paragraph{Third Person.}
The prompt describes a hypothetical individual in the third person, instructing the model to respond about the persona (e.g., used by \citet{jia2024decision,aher2023using,durmus2023towards}).
An example for such a prompt would be: \narrowfont{"Think of a Hispanic woman."}

\paragraph{Interview.}
The prompt contains a Q\&A style dialogue between two speakers and the model continues the conversation from the perspective of one of the speakers (e.g., used by \citet{argyle2023out,santurkar2023whose,kwokevaluating}).\footnote{We use the term interview-based prompting to refer solely to the reformatting of sociodemographic information, unlike \citet{park2024generative}, who supplement persona prompts with qualitative interview content.}
An example would be: \\
\narrowfont{"Interviewer: What gender do you identify as? \\
Interviewee: I identify as 'female'."}

\subsection{Demographic Priming} \label{sec:dem_priming}
We further observe that studies use different linguistic cues to signal the demographics of a persona. The choice of such cues may invoke different group representations depending on the contexts with which such descriptors are associated during training. For instance, explicit demographic descriptors may be more likely to trigger stereotypes associated with demographic groups, as they could be more directly tied to group-based characteristics in the training data. We study race/ethnicity and gender, and present an overview of all 15 intersectional groups analyzed in Appendix \ref{app:soc_prompts}.

\paragraph{Name.}
Demographic identity is signaled implicitly through contextual cues, such as names and titles, relying on an LLM’s implicit associations (e.g., \narrowfont{"Ms. Hernandez"}) used in \citet{aher2023using,wang2025large,giorgi2024modeling}, \textit{inter alia}. In our study, we use last names, titles and pronouns to imply demographic groups.\footnote{cf. Appendix \ref{app:soc_prompts} for the full list.}
\paragraph{Explicit.}
The persona is introduced using explicit descriptors expressed in natural language (e.g., \narrowfont{"a Hispanic woman"}). Examples of studies using this approach are \citet{cheng2023compost,wright2024revealing,kamruzzaman2024woman}.

\paragraph{Structured.}
The persona is described using both explicit descriptors and category labels. This approach mirrors the format used in structured datasets or surveys, where categories are named and corresponding values are assigned (e.g., \narrowfont{"a person of gender 'female'"}), used in~\citet{beck2024sensitivity,hu2024quantifying,do2025aligning}, \textit{inter alia}. 

\section{Experimental Setup}
Our experiments follow a two-part prompt structure. Each prompt template consists of a \textit{persona} segment, where we vary both the role adoption format and demographic priming, followed by a \textit{task} segment that remains consistent across conditions.  
To ensure that the observed effects can be attributed to the variations introduced in section~\ref{sec:prompt_format}, and not to wording unrelated to these dimensions, we include two alternative phrasings for each template.
We then use them to simulate various demographic personas, covering 15 intersectional race/ethnicity and gender groups (8 groups for the closed-ended task). 
An overview of all prompt templates and the full set of demographic descriptors can be found in Appendix~\ref{app:soc_prompts}. We conduct all analyses in English.

\subsection{Tasks}
We use open- and closed-ended tasks to evaluate how well LLMs represent different demographic groups. Our focus is to investigate the extent to which model responses vary across demographic groups and prompt types,
particularly in contexts where differences between groups might lead to stereotypical biases (i.e., writing self-descriptions and social media bios), as opposed to contexts where variations are explicitly required for alignment (i.e. answering surveys). 

\paragraph{Open-Ended.} \hypertarget{par:open_tasks}{}
To assess how LLMs portray various sociodemographic groups in open-generation settings, we use two tasks: \textbf{Self-Description} and \textbf{Social Media Bio}. In both tasks, the model is instructed to generate text from the perspective of an assigned persona --- either a detailed self-description or a brief biography for a social media platform. For each persona prompt (i.e., combination of prompt template and demographic persona), we sample 100 responses and analyze them for various aspects of demographic bias.
While self-descriptions yield longer, more detailed responses, social media bios enable us to examine if similar patterns manifest in shorter text. For each open-ended task, we create 1,080 prompts, covering 15 demographic groups, 9 prompt types (that is, combinations of role adoption and demographic priming strategies), and 2 alternative phrasings.\footnote{\label{f2_names} For demographic priming with \textit{names}, we prompt with 10 different names per demographic group and collect 10 responses each (see Appendix \ref{app:soc_prompts}).}

\paragraph{Closed-Ended.} \hypertarget{par:closed_tasks}{}

To investigate the impact of sociodemographic steering in LLMs more holistically, we also deploy our prompt strategies to a closed-ended \textbf{Survey Response} task. We use the \textit{OpinionsQA} dataset~\cite{santurkar2023whose}, which is widely used to assess the alignment of LLMs with different demographic groups~\cite[\textit{inter alia}]{suh2025languagemodelfinetuningscaled,dominguez2024questioning,hwang2023aligning}, based on questions from the Pew Research Center's American trends Panel (ATP).~\footnote{\url{https://www.pewresearch.org/the-american-trends-panel/}} We sample 100 English-language questions from the most recent ATP waves (54, 82, and 92) and prompt LLMs to answer these as the assigned persona. Following ~\citet{santurkar2023whose}, each prompt includes a single question with multiple answer options. 
We run each prompt once, extract log probabilities to build opinion distributions for each subgroup, and evaluate the distance between these and the corresponding U.S. demographic opinion distributions. This process yields 57,600 prompts, spanning 8 demographic groups\footnote{The closed-ended task includes fewer demographic groups because the OpinionsQA dataset doesn't cover nonbinary gender identity and Middle Eastern race/ethnicity.}, 9 prompt types, 2 alternative phrasings and 100 questions. 

\subsection{Models}
For generalizability, we use five instruction-tuned LLMs of varying parameter sizes from three different model families. We run our experiments with Llama-3 (\texttt{Llama-3.3-70B-Instruct}, \texttt{Llama-3.1-8B-Instruct})~\cite{grattafiori2024llama}, OLMo-2 (\texttt{OLMo-2-0325-32B-Instruct}, \texttt{OLMo-2-1124-7B-Instruct})~\cite{groeneveld2024olmo} and Gemma-3 (\texttt{gemma-3-27b-it})~\cite{team2024gemma}. All computational details are in Appendix~\ref{app:comp_details}.

\subsection{Evaluation}
We evaluate the results of the two types of tasks using criteria aligned with their specific objectives. 

\subsubsection{Open-Ended Evaluation} \label{sec:open_eval}
The open-ended tasks, i.e., writing self-descriptions or social media bios, are intended to reflect a persona's social identity. However, past research has pointed out that demographic characteristics like race and gender represent only one aspect of social identity, which is much more complex and multifaceted  \citep{holck2016identity}. In fact, there is little evidence to suggest that people’s self-descriptions can be distinguished solely by racial and gender dimensions in real life, especially for marginalized groups \citep{haimson2018social, nguyen2014gender}. As such, \textit{LLMs should avoid differentiating self-descriptions of demographic groups solely based on race or gender}, as doing so risks essentializing these characteristics and tokenizing groups, i.e., linking the construction of a person's identity only to their race and gender.

We first preprocess the generated self-descriptions and social media bios by removing demographic markers repeated from the prompt and excluding instances where the model deviated from its assigned persona and instead behaved as an AI assistant (cf. Appendix \ref{sec:preprocessing} for details). We then analyze the remaining responses for various manifestations of bias and stereotyping.
Because measuring stereotypes is inherently difficult and subjective, we use multiple complementary measures to establish convergent validity and to triangulate both the presence and extent of demographic biases across different prompt types.

\paragraph{Stereotypical Bias.}
Using the Marked Personas framework by~\citet{cheng2023marked}, we identify sets of \textit{marked words}, i.e., terms that occur significantly more often in responses about a marked demographic group compared to an unmarked reference group. This approach is grounded in the sociolinguistic concept of \textit{markedness}, which distinguishes explicitly marked categories from unmarked defaults, reflecting how dominant groups tend to be linguistically unmarked and assumed as the default, while non-dominant groups are marked both linguistically and socially due to their group membership~\citep{waugh1982marked}.
Following~\citet{cheng2023marked}, we treat \textit{White} and \textit{male} as the unmarked reference groups for race and gender, respectively.
As a quantitative measure, we count the \textbf{number of marked words} for each combination of demographic group and prompt type. While not every marked word is tied to a stereotype, higher counts may indicate reduced lexical diversity and greater divergence from the unmarked group.
Furthermore, we train a one-vs-all SVM classifier to distinguish responses of personas associated with different demographic groups, following the implementation and response-anonymization procedure of \citet{cheng2023marked}. Higher  \textbf{accuracy} of this classifier suggests that responses for a given group are more linguistically distinct, implying stronger demographic-specific patterns.

\paragraph{Semantic Diversity.}
Building on~\citet{wang2025large}, who found that LLMs produce flatter, more homogeneous responses than humans, we emphasize the importance of diversity within groups: responses should reflect the wide range of interests, values, and perspectives naturally present in any demographic group \citep{lee2024large}.
For the two open-ended tasks, we calculate the semantic diversity of responses based on sentence embeddings. After applying the redaction steps described in Appendix~\ref{sec:preprocessing}, we embed all open-ended responses using \texttt{intfloat/multilingual-e5-large-instruct}, a state-of-the-art multilingual embedding model according to the Massive Text Embedding Benchmark (MTEB) leaderboard~\citep{enevoldsen2025mmtebmassivemultilingualtext}. We form clusters based on each unique combination of prompt type and demographic group, before calculating the \textbf{mean pairwise distance} of all embedding vectors in each cluster. Higher pairwise distances suggest greater semantic diversity, which is desirable, as it may reflect a broader range of perspectives within a demographic group.

\paragraph{Language Switching.}
We classify the language of each open-ended response using \texttt{lingua}.\footnote{\url{https://github.com/pemistahl/lingua-py}} We observe many code-mixed responses with hard to delineate substrings---e.g., the following bio: \narrowfont{"Dominicana | Coffee, libros, y buena vibra | Mamá to a wild one | Sharing life one cafecito at a time"}---, so we only keep the most probable detected language for each response. 
Since our prompts are fully written in English, we would expect all responses to be in English as well, and interpret any kind of language switching as a form of \textit{perpetual foreigner stereotype}~\cite{dennis2018exploring}.

\subsubsection{Closed-Ended Evaluation} \label{sec:eval_align}
In contrast to the open-ended tasks, variation across demographic groups in the closed-ended Survey Response task is not only acceptable but expected. For this task, differences between groups should reflect empirically grounded variations in opinions. 

\paragraph{Opinion Distance.}
Following~\citet{santurkar2023whose}, we extract the log probabilities of the models' answers to survey questions and convert them to answer distributions that can be compared to human answer distributions for the same question. 
We compute distributions separately for each demographic group and prompt type, and quantify opinion distance using the \textbf{average Wasserstein distance} between model and human distributions, where lower values indicate lower opinion distance (i.e., better distributional alignment) ~\citep{suh2025languagemodelfinetuningscaled}. 
As a baseline, we compute the opinion distance between humans and random answers. We also conduct a robustness check by comparing answer distributions from LLM log probabilities with those from verbalized LLM responses across multiple runs in Appendix~\ref{app:val_logprobs}.

\section{Results} \label{sec:results}

We now report the results of our analyses of the five different LLMs by first focusing on \textbf{(i) Demographic Representativeness}, i.e., how well the LLMs simulate different identity groups for the two types of tasks and their corresponding evaluation metrics. Then, we assess the impact of \textbf{(ii) Strategies for Persona Prompting}. Finally, we compare the five different models and explore whether some are more representative than others. For the open-ended tasks, we show results for the Self-Description task in the main body and report complementary results for the Social Media Bio task in Appendix \ref{app:bio}.

\subsection{Demographic Representativeness}
\paragraph{Is there a difference in LLMs' ability to simulate different demographic groups?}

\begin{figure*}[htbp]
    \centering
    \begin{subfigure}[b]{0.49\textwidth}
        \centering
        \includegraphics[width=\linewidth]{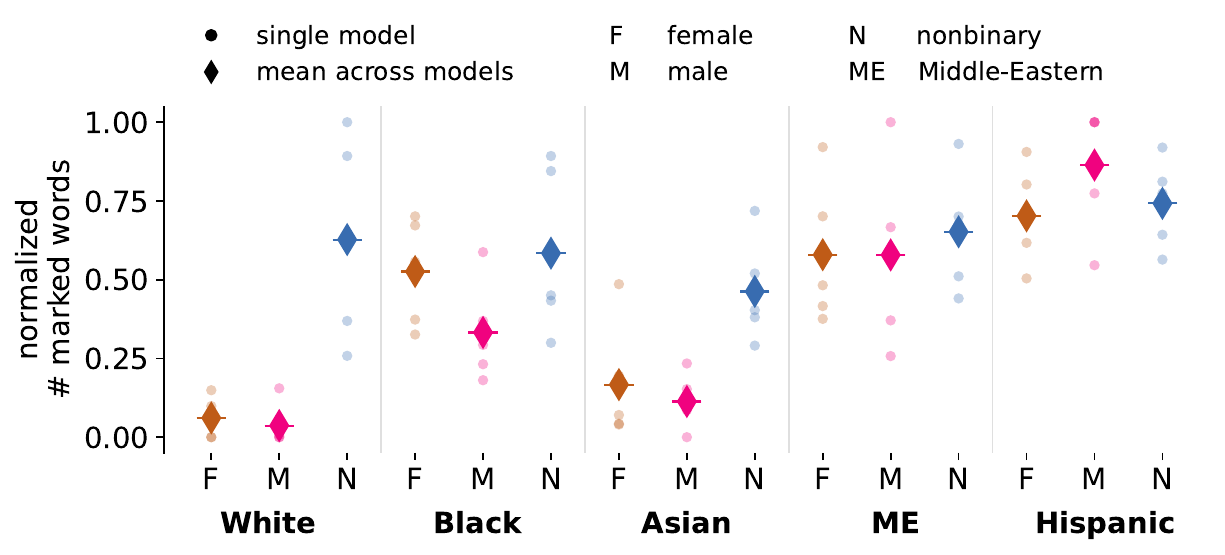} 
        \caption{normalized marked word count ($\downarrow$)}
        \label{fig:sub_mw_sd}
    \end{subfigure}
    \hfill
    \begin{subfigure}[b]{0.49\textwidth}
        \centering
        \includegraphics[width=\linewidth]{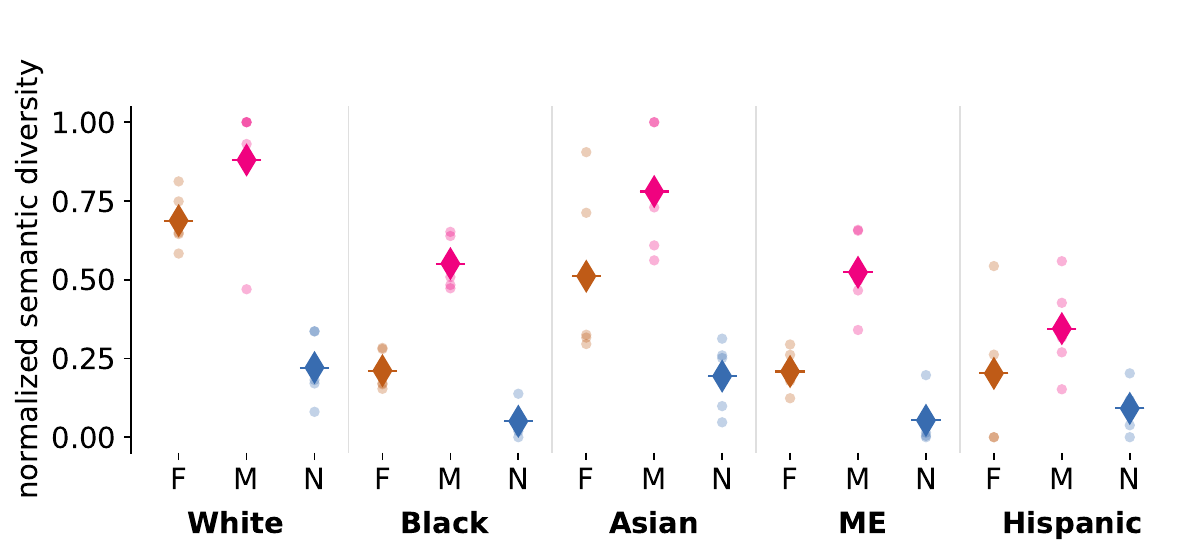}
        \caption{normalized semantic diversity  ($\uparrow$)}
        \label{fig:sub_sdiv_sd}
    \end{subfigure}
    \caption{\textbf{Discrepancies in demographic group representation.} We find systematic differences in self-descriptions of simulated demographic personas. We show the (a) number of marked words and (b) semantic diversity of generated self-descriptions for each demographic group. Values are aggregated across all prompt types and we apply min-max normalization for each model separately to indicate the relative ranking of groups. We observe that self-descriptions for \textit{nonbinary} (N) personas generally exhibit the least favorable outcome (i.e., high marked word count and low semantic diversity), while simulations of \textit{male} (M) personas lead to the most favorable results (i.e., low marked word count and high semantic diversity). Additionally, simulations of \textit{Middle-Eastern (ME)} and \textit{Hispanic} personas are generally associated with less favorable outcomes.
    }
    \label{fig:Sem_div_dem_SD}
\end{figure*}

To assess how LLMs simulate different demographic groups, we first analyze the average number of identified marked words and semantic diversity of demographic groups. 
Figure~\ref{fig:Sem_div_dem_SD} shows that self-descriptions associated with \textit{nonbinary} personas exhibit significantly lower semantic diversity and more marked words, while responses for \textit{male} personas are among the most semantically diverse. 
We further observe that responses associated with \textit{Middle Eastern} and \textit{Hispanic} personas tend to contain more marked words and exhibit lower semantic diversity as opposed to other race/ethnicity groups.~\footnote{We see similar results for social media bios and report those results in Figure~\ref{fig:Sem_div_dem_Bio} in Appendix \ref{app:bio}.}
This suggests that \textbf{simulations of marginalized gender and race/ethnicity groups are more prone to flattened portrayals}, i.e., reduced diversity and recurring, possibly stereotypical patterns.

Next, we examine whether LLMs are prone to language switching during persona simulation.
Across all LLMs, we find that non-English self-descriptions and bios are most frequent for \textit{Hispanic} personas (up to $10.5\%$), while non-English responses are rare for all other racial groups (up to $0.8\%$). This indicates that LLMs fall victim to the \textbf{`perpetual foreigner stereotype' when simulating \textit{Hispanic} personas}~\citep{dennis2018exploring}.

\begin{figure}[tbp]
    \centering
    \includegraphics[width=\linewidth]{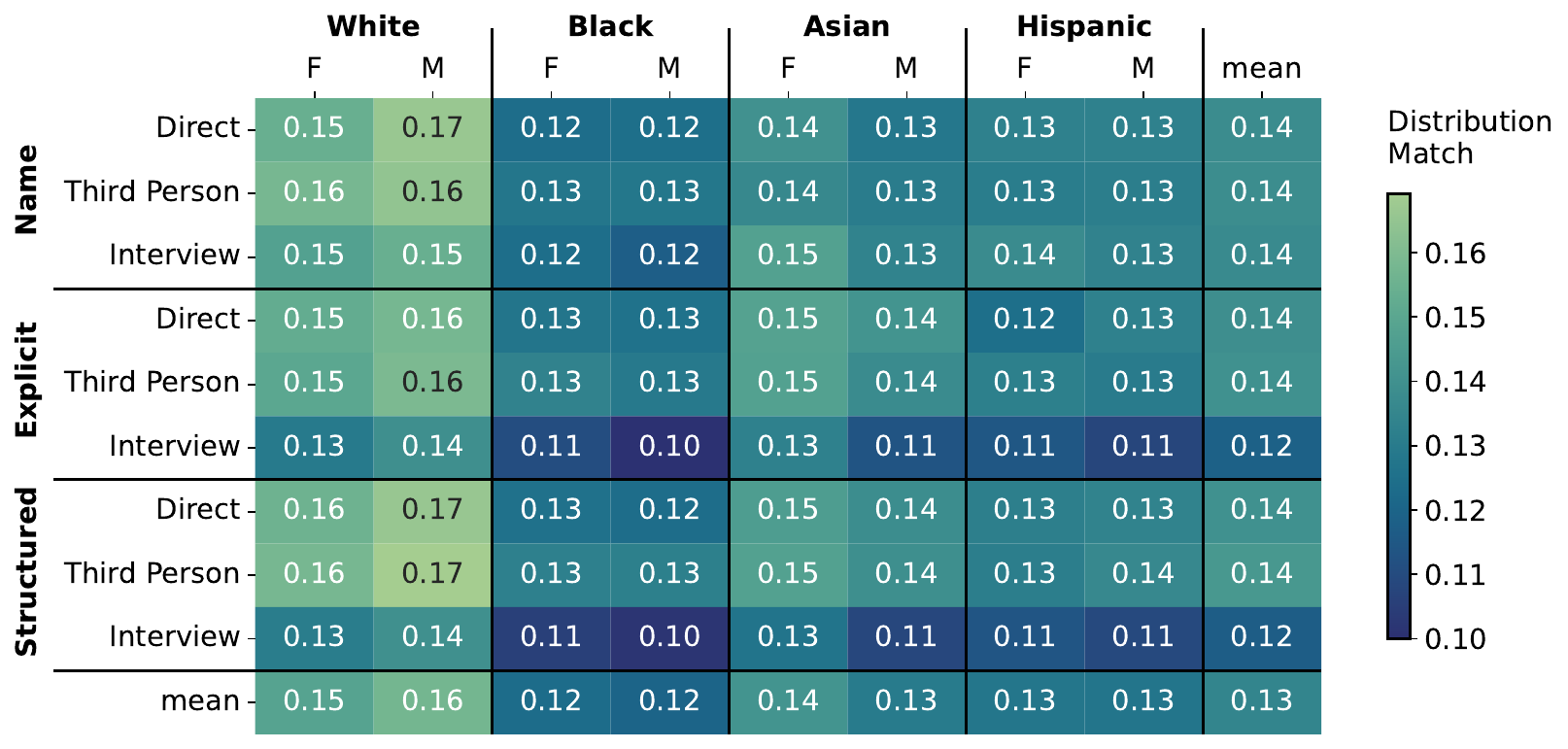}  
    \caption{ \textbf{Opinion distance on OpinionsQA ($\downarrow$).} Abbreviations: M = \textit{male}, F = \textit{female}. We report the average Wasserstein distance for the best-performing model, OLMo-2-7B. Differences across prompt types are generally modest, but the \textit{interview} format leads to improved opinion distance (i.e., lower Wasserstein distance). We show the remaining models in Fig.~\ref{fig:QA_dist} in Appendix~\ref{app:OpinionQA}.}
    \label{fig:opinionqa_olmo7}
\end{figure}

Finally, we compare the opinion distance between simulated personas and the human ground-truth on OpinionsQA (Fig.~\ref{fig:opinionqa_olmo7}). In line with  \citet{suh2025languagemodelfinetuningscaled}, we observe the lowest distance for \textit{Black} personas and the highest distance for \textit{White} personas.
However, we also find that two of the five models yield greater opinion distance than the random baseline (0.25 $\pm$ 0.002). We discuss this in more detail in Section \ref{par:model_comp}.

\subsection{Strategies for Persona Prompting}
\paragraph{How do different prompt types impact LLMs' simulation abilities?}

\begin{figure*}[htbp]
    \centering
    \begin{subfigure}[b]{0.49\textwidth}
        \centering
        \includegraphics[width=\linewidth]{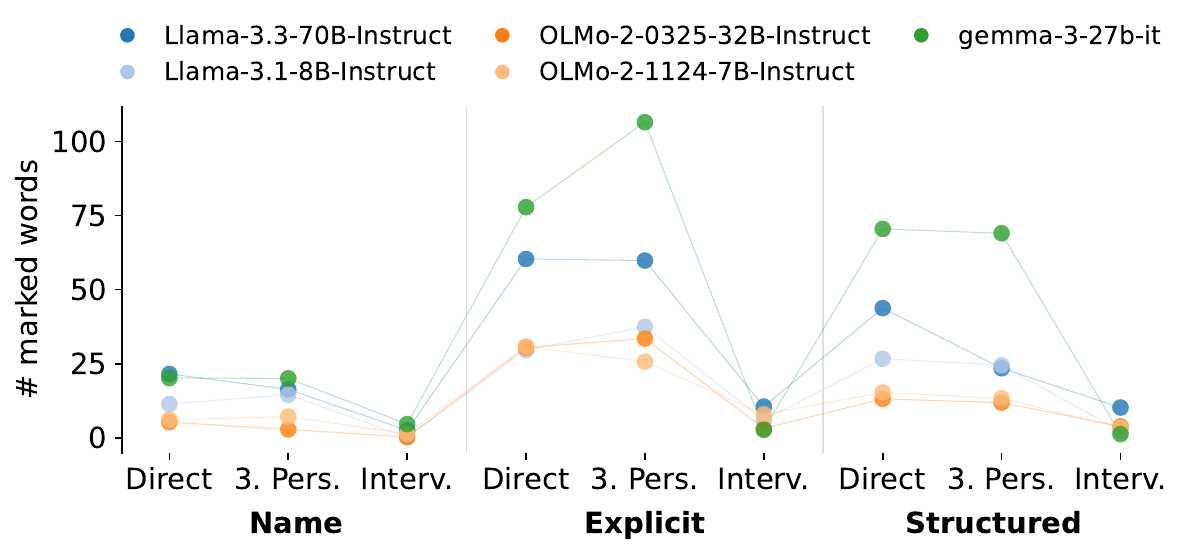}
        \caption{marked word count ($\downarrow$)}
        \label{fig:marked_words_model}
    \end{subfigure}
    \hfill
    \begin{subfigure}[b]{0.49\textwidth}
        \centering
        \includegraphics[width=\linewidth]{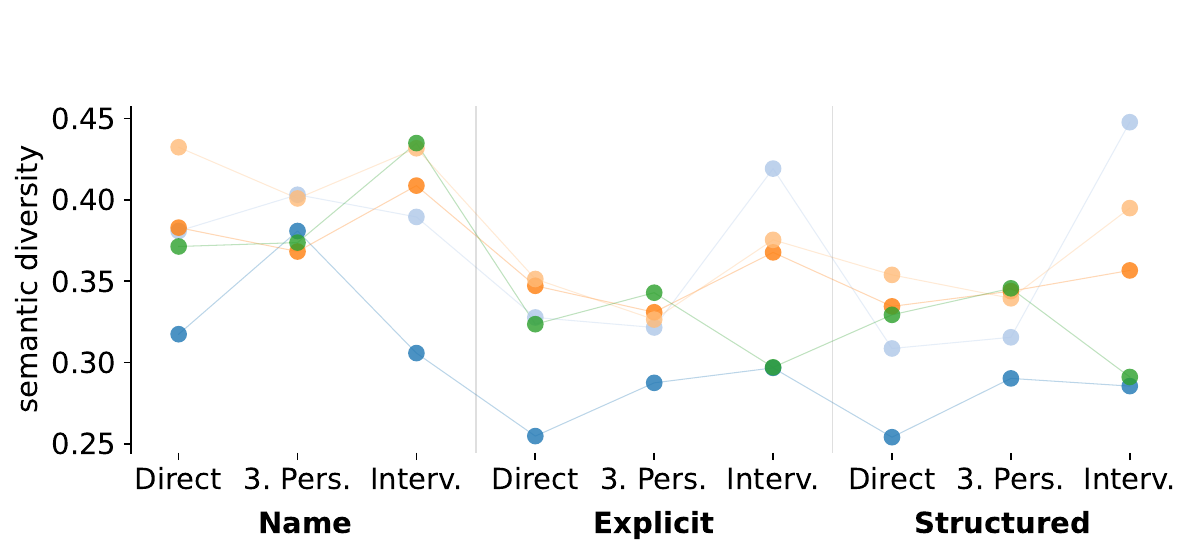}
        \caption{semantic diversity ($\uparrow$)}
        \label{fig:sem_div_model}
    \end{subfigure}
    \caption{ \textbf{Comparison of prompt types and models.} We present the (a) number of marked words and (b) semantic diversity of simulated self-descriptions for each prompt type and model. Values are aggregated across all demographic groups. We find that prompting with \textit{names} and using the \textit{interview} format leads to a lower (i.e., better) marked word count for all models. We observe a similar pattern for semantic diversity, with the exception of Gemma-3-27b and Llama-3.3-70B, which generally exhibit the worst performance across both measures (i.e., high marked word count and low semantic diversity). 
    }
    \label{fig:MW_SD}
\end{figure*}

We analyze prompt types using the Marked Personas framework and find that prompts incorporating \textit{names} for demographic priming and the \textit{interview} format for role adoption produce the fewest marked words in self-descriptions across all models (Fig.~\ref{fig:marked_words_model}).
Beyond the Marked Personas framework, we find that prompts using \textit{names} and the \textit{interview} format also promote overall higher semantic diversity (Fig.~\ref{fig:sem_div_model}).

\begin{figure}[tbp]
    \centering
    \includegraphics[width=\linewidth]{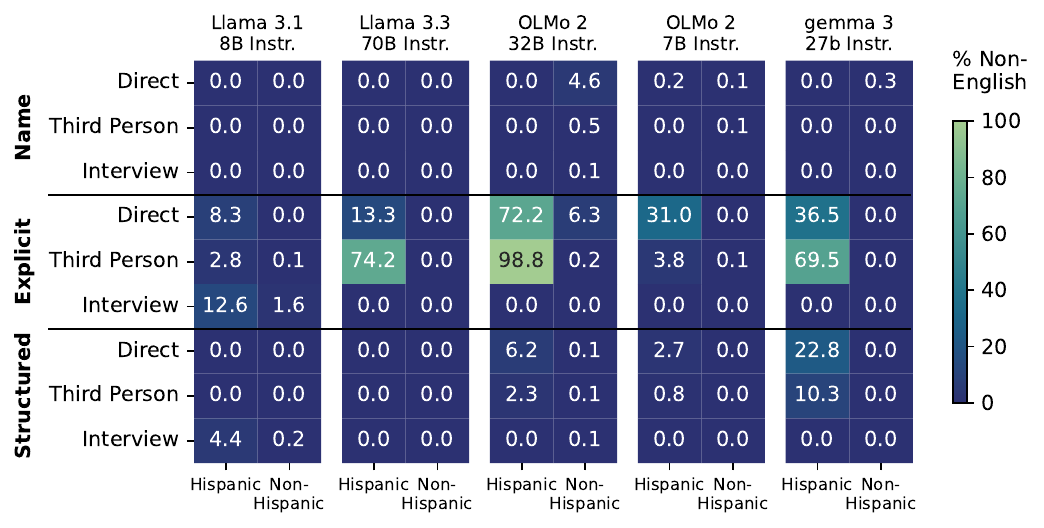}
    \caption{\textbf{Percentage of non-English self-descriptions.} We report the percentage of non-English responses generated for \textit{Hispanic} personas, who receive the highest proportion of such responses. \textit{Explicit} demographic priming leads to higher rates of non-English responses. 
    }
    \label{fig:fig_non_english_SD}
\end{figure}

Figure~\ref{fig:fig_non_english_SD} further shows that prompts using \textit{names} and the \textit{interview} format lead to near-zero rates of non-English responses. Notably, the \textit{interview} format seems to mitigate the language-switching effect introduced by \textit{explicit} demographic priming.~\footnote{The reported patterns, specifically that \textit{names} and the \textit{interview} format lead to to improvements for the open-text measures, also hold for classification accuracy (see Fig.~\ref{fig:acc_all_tasks} in Appendix \ref{app:marked_words}) and for the Social Media Bio task, with detailed results in Figures~\ref{fig:MW_Sem_div_Bio} and \ref{fig:lang_bio} in Appendix \ref{app:bio}.}

Finally, Figure~\ref{fig:opinionqa_olmo7} shows the opinion distance in OpinionsQA (i.e., average Wasserstein distance) for the best performing model, OLMo-2-7B.
We note a performance gain (i.e., lower Wasserstein distance) when using the \textit{interview} format, indicating significantly reduced opinion distance.

Our findings indicate that \textbf{using \textit{names} for demographic priming and/or the \textit{interview} format for role adoption leads to better representation of demographic groups across all models.}\footnote{Further robustness checks and statistical significance for these results can be found in Appendix \ref{app:regression}.} 

To assess whether the observed positive effects benefit all groups equally, we compute the standard deviation of each measure across demographic groups. For each LLM, we run OLS regressions with this standard deviation as the dependent variable, analyzing the effect of prompt types. Table~\ref{tab:reg_mw_std} reports the results for the marked word count, showing that prompts using \textit{names} and the \textit{interview} format significantly reduce the standard deviation across groups, indicating reduced disparities between demographic groups.
This generalizes across all open-text measures for OLMo-2-32B and OLMo-2-7B, indicating that these prompt types improve both overall representation and reduce inter-group disparities. Effects for other models are more mixed, suggesting that those strategies sometimes benefit certain groups more than others. We report regression results for the remaining measures in Table \ref{tab:std} in Appendix \ref{app:var_dem}.
 
\newcolumntype{d}[1]{S[table-format=#1]}
\begin{table}[htb]
    \centering
    \scriptsize
    \renewcommand{\arraystretch}{0.8}
    \setlength{\tabcolsep}{2pt}  
    \begin{tabular}{l d{-1.1} d{-1.1} d{-1.1} d{-1.1} d{-2.1}}
    \toprule
     & \multicolumn{1}{c}{Llama-70B} & \multicolumn{1}{c}{Llama-8B} & \multicolumn{1}{c}{OLMo-32B} & \multicolumn{1}{c}{OLMo-7B} & \multicolumn{1}{c}{Gemma-27b} \\
    \midrule
    \textcolor{okabe_orange}{Name} & -5.9* & -3.2* & -7.3* & -3.5* & -11.4* \\
    \textcolor{okabe_orange}{Struct.} & -3.7* & -1.5* & -6.3* & -2.4* & -6.7 \\
    \textcolor{okabe_red}{Interview} & -5.3* & -2.9* & -4.9* & -2.6* & -10.0* \\
    \textcolor{okabe_red}{3. Person} & -0.4 & 0.8 & -0.9 & -0.6 & -0.3 \\
    \midrule
    {Self-Descr.} & 6.3* & 3.1* & 5.7* & 3.7* & 12.4* \\
    Phrasing v2 & 0.6 & 0.2 & 0.7 & 1.0 & 0.8 \\
    \midrule
    Intercept & 8.0* & 4.1* & 7.0* & 3.2* & 12.3* \\
    \bottomrule
    \end{tabular}
    
    \caption{
    \textbf{Modeling group disparities in marked word counts.} 
    We conduct OLS regression analyses per LLM using the standard deviation of the marked word count between demographic groups as a dependent variable and report the regression coefficients. The independent variables include: \textcolor{okabe_orange}{demographic priming} (reference: explicit), \textcolor{okabe_red}{role adoption} (reference: direct), {task} (reference: Bio), and prompt phrasing (reference: v1). Lower standard deviation~($\downarrow$) indicates reduced disparities between demographic groups. We find that using \textit{names} and the \textit{interview} format significantly reduces disparities in marked word counts across all models. *~$p < 0.05$.}
    \label{tab:reg_mw_std}
\end{table}

\subsection{Model Comparison} \label{par:model_comp}
Surprisingly, two of the largest models, i.e., Llama-3.3-70B and Gemma-3-27B, perform worst with respect to the Marked Personas framework and semantic diversity, while the smaller OLMo-2-7B, performs best (Fig. \ref{fig:MW_SD}).~\footnote{We observe the same pattern for classification accuracy (see Fig.~\ref{fig:acc_all_tasks} in Appendix \ref{app:marked_words}) and for the Social Media Bio task (see Fig. \ref{fig:MW_Sem_div_Bio} in Appendix \ref{app:bio}).}

Both Llama-3.3-70B and Gemma-3-27B also fail to match human answer distributions on OpinionsQA, with average Wasserstein distances worse than the random baseline (cf. Fig. \ref{fig:QA_dist} in Appendix \ref{app:OpinionQA}). Further analysis reveals highly skewed log probabilities; for example in 88\% of instances, LLama-3.3-70B assigns a probability higher than 0.999 to a single answer option. 
Since human distributions are rarely this extreme, this likely causes the high overall Wasserstein distance. Nonetheless, Llama-3.3-70B performs well in finding the overall majority opinion (cf. Figure \ref{fig:opinionqa_llama70} in Appendix \ref{app:OpinionQA}), indicating that the model is familiar with the overall preference of groups, but is unable to model a human-like preference distribution. Our findings are in line with~\citet{suh2025languagemodelfinetuningscaled}, and point to the need for further investigation of the log probabilities of large(r) instruction-tuned models. Overall, our results 
indicate that \textbf{larger models are not necessarily better, but can actually be less representative for some demographic groups.}

\begin{table*}[tb]
    \centering
    \small
    \begin{tabular}{ll}
        \toprule
        Group & Top-10 Marked Words \\
        \midrule
        Asian woman & \textcolor{olive}{heritage}, dark, petite, long, modern, almondshaped, \textcolor{olive}{cultural}, bun, golden, delicate \\
        Asian man & chinese, korean, respect, styled, \textcolor{olive}{heritage}, martial, vietnamese, slender, japanese, \textcolor{olive}{traditional} \\
        Asian nonbinary pers. & \textcolor{magenta}{identity}, \textcolor{magenta}{gender}, \textcolor{olive}{traditional}, \textcolor{magenta}{binary}, art, \textcolor{olive}{cultural}, \textcolor{olive}{heritage}, embracing, blend, \textcolor{magenta}{pronouns} \\
        Black woman & african, sister, \textcolor{darkred}{strength}, daughter, \textcolor{darkred}{resilience}, \textcolor{darkred}{resilient}, rich, justice, unapologetically, ancestors \\
        Black man & african, son, \textcolor{darkred}{resilience}, brother, community, rich, \textcolor{darkred}{strength}, \textcolor{olive}{proud}, justice, positive \\
        Black nonbinary pers. & \textcolor{magenta}{identity}, \textcolor{magenta}{gender}, african, fluidity, \textcolor{darkred}{resilience}, \textcolor{magenta}{binary}, world, art, blackness, unique \\
        Hispanic woman & latina, vibrant, \textcolor{olive}{proud}, spanish, \textcolor{olive}{heritage}, warm, american, latin, mujer, mi \\
        Hispanic man & latin, \textcolor{olive}{proud}, american, spanish, family, la, salsa, mi, hombre, que \\
        Hispanic nonbinary pers. & \textcolor{magenta}{identity}, latinx, \textcolor{olive}{heritage}, \textcolor{magenta}{gender}, latin, vibrant, justice, \textcolor{magenta}{binary}, social, puerto \\
        ME woman & rich, \textcolor{olive}{heritage}, \textcolor{dodgerblue}{faith}, \textcolor{olive}{traditions}, women, warmth, modernity, deeply, independent, \textcolor{olive}{tradition} \\
        ME man & hospitality, \textcolor{dodgerblue}{faith}, arabic, region, ancient, rich, \textcolor{olive}{heritage}, strong, \textcolor{dodgerblue}{muslim}, middle \\
        ME nonbinary pers. & \textcolor{magenta}{identity}, \textcolor{olive}{heritage}, \textcolor{olive}{traditional}, \textcolor{magenta}{gender}, middle, tapestry, \textcolor{olive}{cultural}, east, rich, \textcolor{magenta}{binary} \\
        White woman & blonde, long, wavy, yoga, slender, curly, approachable, fair, olive, blue \\
        White man & short, outdoor, feet, straightforward, blue, starting, stubble, computer, beer, honesty \\
        White nonbinary pers. & \textcolor{magenta}{gender}, \textcolor{magenta}{binary}, \textcolor{magenta}{identity}, \textcolor{magenta}{pronouns}, \textcolor{olive}{traditional}, use, androgynous, categories, fluid, creative \\
        \bottomrule
    \end{tabular}
    \caption{\textbf{Top-10 marked words in self-descriptions by demographic group.} 
    For each group, we report the 10 most frequent marked words aggregated across prompt types. We highlight words linked to problematic stereotypes (see \citet{cheng2023marked}); specifically (1) the narrative of imposed resilience for \textit{Black} personas (\textcolor{darkred}{red}), (2) the relationship to demographic identity as the primary frame for \textit{non-White} groups (\textcolor{olive}{olive}), (3) the conflation of Middle-Eastern (ME) identities with religiosity (\textcolor{dodgerblue}{blue}), (4) a disproportionate focus on gender identity for \textit{nonbinary} personas (\textcolor{magenta}{magenta}).}
    \label{tab:top_words}
\end{table*}

\section{Analysis of Marked Words} \label{sec:analysis_of_mw}
To complement the quantitative findings in Section \ref{sec:results}, we also analyze the marked words identified in self-descriptions across all combinations of demographic group and prompt type.
To this end, we compute the most frequent marked words per demographic group and present the top 10 in Table \ref{tab:top_words}. In line with \citet{cheng2023marked}, we observe that many of these words reflect patterns of markedness, essentialism, and othering in LLM responses. In particular, we identify four categories of marked words linked to problematic stereotypes as detailed in their work:
\begin{enumerate}[itemsep=0.3em, topsep=0.3em, parsep=0em, partopsep=0em]
    \item Narrative of imposed resilience for \textit{Black} personas (e.g., \textcolor{darkred}{``resilience''}, \textcolor{darkred}{``strength''})
    \item Relationship to demographic identity as the primary frame of any \textit{non-White} groups (e.g., \textcolor{olive}{``heritage''}, \textcolor{olive}{``culture''})
    \item Conflation of \textit{Middle-Eastern} identity with religiosity (e.g., \textcolor{dodgerblue}{``faith''}, \textcolor{dodgerblue}{``muslim''})
    \item Disproportionate focus on gender identity for \textit{nonbinary} personas (e.g., \textcolor{magenta}{``gender''}, \textcolor{magenta}{``identity''})
\end{enumerate}

\begin{figure*}
    \centering
    \begin{subfigure}[b]{0.49\textwidth}
    \centering
        \includegraphics[width=\linewidth]{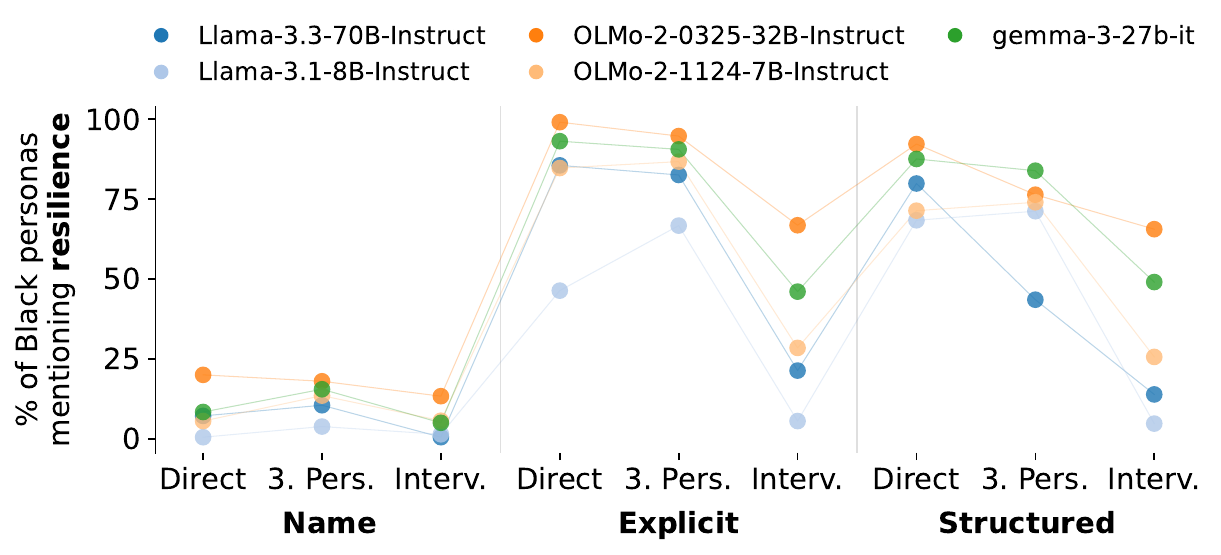}  
        \caption{narrative of imposed resilience}
    \end{subfigure}
    \hfill
    \begin{subfigure}[b]{0.49\textwidth}
    \centering
        \includegraphics[width=\linewidth]{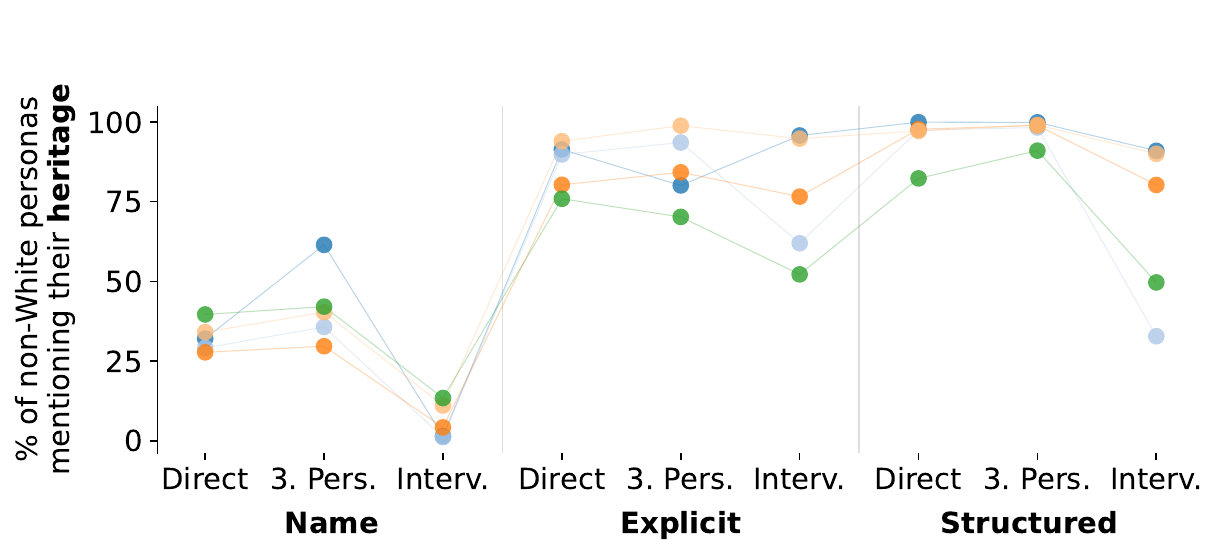}  
        \caption{relationship to dem. identity as primary frame}
    \end{subfigure}

    \caption{\textbf{Share of stereotyped self-descriptions across prompt types.} We report the share of self-descriptions including terms (a) associated with the imposed resilience narrative (e.g., ``resilience'', ``strength'') for \textit{Black} personas, and (b) emphasizing their relationship to demographic identity (e.g., ``heritage'', ``culture'') for \textit{non-White} personas.} 
    \label{fig:resilience}
\end{figure*}

For each category, we calculate the share of personas from the associated groups whose self-descriptions include such words and compare between prompt strategies. Figure \ref{fig:resilience} shows results for category \textcolor{darkred}{1} and \textcolor{olive}{2}: using \textit{names} and/or the \textit{interview} format consistently lowers the share of stereotyped persona descriptions across models.\footnote{Results for additional stereotype categories follow the same pattern; full results and significance levels are provided in Appendix \ref{app:ster_cat}.} 

\section{Related Work}
\textbf{Personas in LLMs.} 
Personas, widely used in Interaction Design to represent user archetypes~\cite{cooper2014face}, have recently been adapted to mold the behavior of language technologies like chatbots and LLMs.
Personas in LLMs span e.g. psychometric, occupational, and demographic dimensions to guide model behavior and tailor responses to contexts or user needs~\cite{tseng2024two}.
Inducing personas in LLMs effectively is an active area of research~\cite{jiang2023evaluating,shu2024you,liu2024evaluating}. We study demographic personas which can be used to create synthetic samples of particular human subpopulations~\cite{argyle2023out}. 

\textbf{Demographic Biases in LLMs. }There is extensive research on demographic biases in LLM; see e.g.,~\citet{gupta2023sociodemographic,sen_2025} for overviews. 
Prior work shows that LLMs associate names with identity groups~\cite{gupta2023bias,pawar2025presumed} or respond to users differently based on cultural perceptions~\cite{dammu2024they}.

Biases in LLMs can also crop up when simulating or portraying different groups of people, causing representational harms. Some research suggests that demographic biases in LLMs may help align their outputs with the views of specific groups~\cite{argyle2023out}. To assess representativeness, recent work has employed survey-based 
measures~\cite{santurkar2023whose,atari2023humans}, reporting mixed results across demographic groups. Other studies have used alternative methods such as content analysis, free-text self-descriptions, and behavioral simulations~\cite{cheng2023marked,cheng2023compost}. Some studies point out that persona-steered LLMs do not always match human behavior~\cite{beck2024sensitivity,hu2024quantifying}. Furthermore, LLMs have also been shown to flatten or reduce the diversity of certain groups, particularly marginalized people, when portraying them~\cite{wang2025large}.
We examine how LLMs represent intersectional racial and gender groups based on sociodemographic steering, using free-text and survey questions.

\section{Discussion and Conclusion}

Our findings show that LLMs often stereotype marginalized groups, especially \textit{nonbinary}, \textit{Hispanic}, and \textit{Middle Eastern} personas, highlighting ongoing challenges in representing demographic groups \citep{cheng2023compost, cheng2023marked}.  

However, we find that the choice of demographic priming and role adoption strategies significantly impacts how these groups are portrayed. 
In particular, \textit{interview}-style prompts can help mitigate stereotypical outputs, enhance diversity, and improve alignment. Based on this, we recommend that practitioners consider this strategy when designing sociodemographic persona prompts. 

Using \textit{last names} for demographic priming also shows promise in mitigating biases and aligns with \citet{wang2025large}, who found similar effects using first names.
However, using first names to signal gender identity, can entrench biases towards certain groups, e.g, nonbinary people~\cite{gautam-etal-2024-stop}. We circumvent this by instead relying on titles (i.e, Mr., Ms., and Mx.) to signal gender. Nevertheless, we observe that models occasionally interpret last names in unintended ways (for example, some names associated with Black identity in the U.S. were linked to French residents), posing challenges to validity \citep{gautam-etal-2024-stop}. Thus, we suggest treating names as a potentially useful but ethically fraught tool, that requires careful implementation and critical reflection on its limitations.

A new observation is that LLMs switch language during simulation, but only when simulating Hispanic personas. To the best of our knowledge, this has not been studied before. 
Investigating whether this extends to other identities and languages, and how it relates to LLMs' multilingual capabilities, is a promising direction for future work.

Finally, we stress the importance of clearly documenting and justifying choices around role adoption, demographic priming, and the specific descriptors used in persona prompts. 
To support future work, we release our code and datasets.\footnote{\url{https://github.com/dess-mannheim/prompt-makes-the-persona}}

\section*{Limitations}

While our work lays a foundation for evaluating sociodemographic persona prompts, several limitations remain. First, the study is limited to English and focuses only on two commonly studied dimensions --- gender and race \citep{sen_2025} --- which constrains generalizability to other demographic dimensions and languages. 
Additionally, because the persona prompts are constructed in a fully data-driven manner, the resulting set is not exhaustive and may miss other meaningful variations. This limitation also extends to the choice of demographic descriptors; for example, we do not systematically compare other demographic descriptors such as \narrowfont{"girl"} or \narrowfont{"father"}, leaving such analysis for future work.

To assess the alignment with human responses on OpinionsQA, we prompt models using a multiple-choice format, reflecting the design of the original human survey. While this has become a standard evaluation paradigm in the field~\citep{hwang2023aligning, moon-etal-2024-virtual}, recent research has highlighted important limitations of this approach~\citep{rottger-etal-2024-political}, which also apply to our work. Additionally, we do not randomize or control the order of answer options due to computational constraints. This may influence model response patterns \citep{rupprecht2025prompt} and, consequently, the observed alignment behavior. Lastly, while we considered the possibility of data contamination in OpinionsQA, we find it unlikely to be a confounding factor, as none of the models show particularly strong alignment on the task.

Finally, we always place the persona prompts in the user message, as our literature review found this to be the most common approach in prior studies. We anticipate that using the system prompt instead could reduce the frequency of role violations, particularly for Gemma-3-27b, which showed the highest rate of such violations (cf. Appendix \ref{app:role_violations}). A systematic comparison of both strategies is an interesting venue for future work.

\section*{Ethical Considerations}
Our work highlights how sociodemographic prompts can improve representation in tasks like writing self-descriptions and answering survey questions. 
The former use-case is an assistive task, one where an LLM user might co-create self-descriptions or even descriptions of other people with the LLM. Therefore, it is imperative that LLMs do not treat different groups solely based on demographic characteristics. The second use case, answering survey questions, is an active area of study in both NLP~\cite{santurkar2023whose,salecha2024large,tjuatja2024llms,dominguez2024questioning} and survey methodology~\cite{argyle2023out,rothschild2024opportunities}. While survey questions can help us measure LLMs' alignment on human behavior and opinions, some researchers claim that human respondents could potentially be substituted by demographically faithful LLMs to fill out surveys on their behalf. However several researchers caution against this usage, both due to methodological challenges~\cite{wang2025large} as well as epistemological questions~\cite{agnew2024illusion}. Therefore, we do not advocate for sociodemographic steering to replace human survey respondents, even if certain prompt styles lead to better results. We also caution against over reliance on quantitative measures alone for measuring LLM's representativeness. Measures such as semantic diversity or the frequency of marked words offer valuable insights, but should not be interpreted in isolation. Instead, they should be interpreted along qualitative analysis of outputs, task context, and the underlying normative goals.

We also acknowledge that our demographic coverage is limited; however, our findings on the \textit{role adoption} format are adaptable to other categories, e.g., political leaning or education. We aimed to include marginalized groups in our study and indeed find that they are susceptible to stereotyping. 

\paragraph{Names as Demographic Proxies.}
Name-based prompting presents additional challenges. The racial categories and last names we use in this paper are based on U.S. Census data, biasing the study toward U.S.-centric naming conventions and limiting applicability to other geographic and cultural contexts. Moreover, focusing only on the most common last names risks capturing a narrow and potentially unrepresentative subset of a group. We further find that LLMs can interpret last names in unintended ways. For instance, last names associated with Black identity in the U.S. were occasionally interpreted as French residents (e.g., \narrowfont{“Mr. Jeanbaptiste, a 35-year-old history enthusiast from Paris”}), indicating that name-based priming may invoke unintended geographic or cultural associations.
Crucially, we caution against relying on \textit{first names} to infer or encode gender, as this risks misgendering individuals and excluding nonbinary and gender-diverse people; therefore, we use titles instead to signal gender. We refer readers to \citet{gautam-etal-2024-stop} for a comprehensive discussion of the limitations and ethical considerations involved in using personal names to infer or signal sociodemographic attributes.

\section*{Acknowledgments}
We thank Florian Lemmerich for the valuable input and helpful discussions in the early stages of this project. The authors acknowledge support by the state of Baden-Württemberg through bwHPC and the German Research Foundation (DFG) through grant INST 35/1597-1 FUGG.

\bibliography{acl_latex}

\appendix

\label{sec:appendix}

\section{Review of Persona Prompt Categories}\label{app:lit_review}
We review a sample of 47 papers that simulate personas in generative LLMs, as categorized by~\citet{sen_2025}. Of these, 44 papers use persona prompts in natural language, while the remaining 3 employ alternative methods such as fine-tuning or soft prompting with persona tokens.
Our analysis reveals substantial variation in how persona prompts are constructed. Specifically, we identify two primary dimensions: \textit{role adoption}, which refers to how the model is instructed to adopt a persona, and \textit{demographic priming}, which concerns how demographic cues are presented to the LLM within the prompt.
For each of these dimensions, we define categories that capture common prompting strategies. We focus primarily on the most prevalent categories observed across the surveyed papers. In the following, we provide usage statistics for these categories and list additional, less common categories that were identified but not used in our main analysis.
Note that several papers used multiple prompts, and are thus counted in more than one category where applicable.
\paragraph{Role Adoption.}
In terms of role adoption, we find that the majority of prompts fall into the three main categories (cf. Section \ref{sec:role_adoption}): \textit{direct} (30 papers), \textit{third person} (14 papers), and \textit{interview} (4 papers). 
We also identified a smaller group of papers (4 in total) that adopted roles using first-person prompts (e.g., \narrowfont{“I am [persona]...”}) within text-completion tasks.
We exclude this category, as we focus on instruction-tuned models, while this is a text-completion prompt.
\paragraph{Demographic Priming.}
For demographic priming, we observe three primary strategies (cf. Section \ref{sec:dem_priming}): \textit{explicit} (26 papers), \textit{structured} (12 papers), and \textit{name} (4 papers).
Additionally, 4 papers combined explicit and structured priming (e.g., \narrowfont{"You are a White individual with a Female gender identity... "}) and 1 paper used prompting language as a proxy for nationality. As language is not equivalent to race/ethnicity, we exclude linguistic prompting from our analysis. Another paper did not specify how demographic priming was conducted.

\section{Sociodemographic Prompt Construction} \label{app:soc_prompts}
In the following section, we list all components used to construct the sociodemographic persona prompts. Note that these prompts are provided in the user prompt, not the system prompt. This decision is based on findings from our literature review, which revealed that most studies adopt this approach. Additionally, access to the system prompt is not always guaranteed, making the user prompt a more applicable option for practitioners.

Table \ref{tab:prompt_templates} presents all prompt templates used in our experiments. 
Task-specific strings are indicated by a placeholder ([TASK]) and substituted as shown in Table \ref{tab:task_templates}.  
Placeholders indicating race/ethnicity through explicit descriptors ([R]) or last names ([NM]) are listed in Table \ref{tab:desc_race}. 
The remaining placeholders are populated with demographic descriptors as indicated in Tables \ref{tab:desc_gender} and \ref{tab:desc_add}.

\paragraph{Last Names.}
We source last names for all racial and ethnic groups from \citet{aher2023using}, except for Middle Eastern names, which are not included in that study. For Middle Eastern race/ethnicity, we instead select the most common last names from the three largest Middle Eastern countries, namely Turkey,\footnote{\url{https://forebears.io/turkey/surnames}} Iran,\footnote{\url{https://forebears.io/iran/surnames}} and Saudi-Arabia.\footnote{\url{https://forebears.io/saudi-arabia/surnames}} For each task, we include all 10 names per demographic group. In the open-ended tasks, each prompt is run with every individual name using 10 random seeds (instead of 100), resulting in 100 responses per demographic category.

All measures are computed on the combined set of responses for all names within the same demographic category (i.e., all names listed in a single row of Table \ref{tab:desc_race}).
For semantic diversity and opinion distance, we additionally calculated the mean pairwise distance of embedding vectors and Wasserstein distance, respectively, for the responses associated with each individual name, and then averaged these values across names within the same demographic category. We found that this alternative approach produced results that were very similar to those obtained from the aggregated data, and, more importantly, the overall patterns remained consistent.

\begin{table*}
    \centering
    \resizebox{0.95\textwidth}{!}{\rotatebox{90}{
    \begin{tabular}{l l p{11cm} p{11cm}}
        \toprule
        Role Adop. & Dem. Priming & Prompt Phrasing \textit{v1}& Prompt Phrasing \textit{v2}\\
        \midrule
        Direct & Explicit & You are a [R] [G1] \textcolor{gray}{who lives in the U.S}. [TASK] & Act as a [R] [G1] \textcolor{gray}{who lives in the U.S}. [TASK] \\
        Direct & Name & You are a person named [P4] [NM] \textcolor{gray}{who lives in the U.S}. [TASK] & Act as a person named [P4] [NM] \textcolor{gray}{who lives in the U.S}. [TASK] \\
        Direct & Structured & You are a person of race/ethnicity '[R]' and gender '[G2]' \textcolor{gray}{who lives in the U.S}. [TASK] & Act as a person of race/ethnicity '[R]' and gender '[G2]' \textcolor{gray}{who lives in the U.S}. [TASK] \\
        Interview & Explicit & You are an interviewee. Based on your previous answers, provide an answer to the last question. \newline
        Interviewer: Do you identify as [G\_list1]? \newline
        Interviewee: I identify as a [G1]. \newline
        Interviewer: Do you consider yourself to be [R\_list1]? \newline
        Interviewee: I consider myself to be [R]. \newline
        \textcolor{gray}{Interviewer: Where do you live? \newline
        Interviewee: I live in the U.S.} \newline
        Interviewer: [TASK] \newline
        Interviewee:... & You are Speaker 2. Based on your previous answers, provide an answer to the last question. \newline
        Speaker 1: Are you [G\_list1]? \newline
        Speaker 2: I am a [G1]. \newline
        Speaker 1: Are you [R\_list1]? \newline
        Speaker 2: I am [R]. \newline
        \textcolor{gray}{Speaker 1: Where do you live? \newline
        Speaker 2: I live in the U.S.} \newline
        Speaker 1: [TASK] \newline
        Speaker 2:... \\
        Interview & Name & You are an interviewee. Based on your previous answers, provide an answer to the last question. \newline
        Interviewer: Please tell me your name. \newline
        Interviewee: My name is [P4] [NM]. \newline
        \textcolor{gray}{Interviewer: Where do you live? \newline
        Interviewee: I live in the U.S.} \newline
        Interviewer: [TASK] \newline
        Interviewee:... & You are Speaker 2. Based on your previous answers, provide an answer to the last question. \newline
        Speaker 1: What is your name? \newline
        Speaker 2: My name is [P4] [NM]. \newline
        \textcolor{gray}{Speaker 1: Where do you live? \newline
        Speaker 2: I live in the U.S.} \newline
        Speaker 1: [TASK] \newline
        Speaker 1:... \\
        Interview & Structured & You are an interviewee. Based on your previous answers, provide an answer to the last question. \newline
        Interviewer: What gender do you identify as? Do you identify as [G\_list2]? \newline
        Interviewee: I identify as '[G2]'. \newline
        Interviewer: What race/ethnicity do you consider yourself to be? Do you consider yourself to be [R\_list2]? \newline
        Interviewee: I consider myself to be '[R]'. \newline
        \textcolor{gray}{Interviewer: Where do you live? \newline
        Interviewee: I live in the U.S.} \newline
        Interviewer: [TASK] \newline
        Interviewee:... & You are Speaker 2. Based on your previous answers, provide an answer to the last question. \newline
        Speaker 1: What is your gender? Are you [G\_list2]? \newline
        Speaker 2: My gender is '[G2]'. \newline
        Speaker 1: What is your race/ethnicity? Are you [R\_list2]? \newline
        Speaker 2: My race/ethnicity is '[R]'. \newline
        \textcolor{gray}{Speaker 1: Where do you live? \newline
        Speaker 2: I live in the U.S.} \newline
        Speaker 1: [TASK] \newline
        Speaker 2:... \\
        Third Person & Explicit & Think of a [R] [G1] \textcolor{gray}{who lives in the U.S}. [TASK] & [TASK] \\
        Third Person & Name & Think of a person named [P4] [NM] \textcolor{gray}{who lives in the U.S}. [TASK] & [TASK] \\
        Third Person & Structured & Think of a person of race/ethnicity '[R]' and gender '[G2]' \textcolor{gray}{who lives in the U.S}. [TASK] & [TASK] \\
        \bottomrule
    \end{tabular}}}
    \caption{\textbf{Overview of our persona prompt templates.} This table presents the persona prompt templates used in our experiments. Text in \textcolor{gray}{gray} is only added for the closed-ended Survey Response task. Tables \ref{tab:task_templates}, \ref{tab:desc_race}, \ref{tab:desc_gender} and \ref{tab:desc_add} specify the possible terms for each placeholder, respectively.}
    \label{tab:prompt_templates}
\end{table*}

\begin{table*}
\centering
\renewcommand{\arraystretch}{1.2} 
    \begin{tabular}{l|p{6cm}| p{6cm}}
    \toprule
         Task &  Explicit and Interview & Third Person \\
    \midrule
         Self-Description & How would you describe yourself? Only respond with the self-description. Use the following JSON format: \{"self\_description": <a detailed self-description>\} &
         How would [P1] describe [P2]? Only respond with the self-description. Use the following JSON format: \{"self\_description": <a detailed self-description>\} \\
         Social Media Bio & What is your social media username and bio? Only respond with the username and bio. Use the following JSON format: \{"username": <social media username>, "bio": <social media bio, without any extra commentary whatsoever>\} &
         What would be [P3] social media username and bio? Only respond with the username and bio. Use the following JSON format: \{"username": <social media username>, "bio": <social media bio, without any extra commentary whatsoever>\}\\
         OpinionsQA & How would you answer the following question: [QUESTION] Only respond with one of the listed answer options. Use the following JSON format: \{"answer\_option": <one of the listed answer options, without any extra commentary whatsoever>\} & 
         How would [P1] answer the following question: [QUESTION]  Only respond with one of the listed answer options. Use the following JSON format: \{"answer\_option": <one of the listed answer options, without any extra commentary whatsoever>\} \\
    \bottomrule
    \end{tabular}
    \caption{\textbf{Task instructions.} This table shows the specific instructions given to the simulated personas for performing each task ([TASK]). The placeholders indicating questions ([QUESTION]) are filled with questions from the OpinionsQA dataset.}    
    \label{tab:task_templates}
\end{table*}

\begin{table*}
\centering
\renewcommand{\arraystretch}{1.2} 
    \begin{tabular}{l|p{12cm}}
    \toprule
        Race/Ethnicity ([R]) & Last Name ([NM])\\
    \midrule
        White & Olson, Snyder, Wagner, Meyer, Schmidt, Ryan, Hansen, Hoffman, Johnston, Larson\\
        Black & Smalls, Jeanbaptiste, Diallo, Kamara, Pierrelouis, Gadson, Jeanlouis, Bah, Desir, Mensah \\
        Asian & Nguyen, Kim, Patel, Tran, Chen, Li, Le, Wang, Yang, Pham \\
        \textcolor{gray}{Middle-Eastern} & \textcolor{gray}{Khan, Ali, Ahmed, Hassan, Yılmaz, Kaya, Demir, Mohammadi, Hosseini, Ahmadi} \\
        Hispanic & Garcia, Rodriguez, Martinez, Hernandez, Lopez, Gonzalez, Perez, Sanchez, Ramirez, Torres \\  
    \bottomrule
    \end{tabular}
    \caption{\textbf{Descriptors for all race/ethnicity groups.} We show all explicit descriptors (used for \textit{explicit} and \textit{structured} demographic priming; [R]) with corresponding last names (for demographic priming with \textit{names}; [NM]). Descriptors in \textcolor{gray}{gray} are only used in the open-ended tasks, since the OpinionsQA dataset for the closed-ended Survey Response task does not cover Middle-Eastern race/ethnicity.}
    \label{tab:desc_race}
\end{table*}

\begin{table*}
\centering
\renewcommand{\arraystretch}{1.2} 
    \begin{tabular}{l|l|l|l|l|l|l}
    \toprule
        Gender & [G1] & [G2] & [P1] & [P2] & [P3] & [P4] \\
    \midrule
        Male (M) & man & male & he & himself & his & Mr. \\
        Female (F) & woman & female & she & herself & her & Ms. \\
        \textcolor{gray}{Nonbinary (N)} & \textcolor{gray}{nonbinary person} & \textcolor{gray}{nonbinary} & \textcolor{gray}{they} & \textcolor{gray}{themselves} & \textcolor{gray}{their} & \textcolor{gray}{Mx.} \\
    \bottomrule
    \end{tabular}
    \caption{\textbf{Descriptors for all gender identities.} We show all descriptors used to signal gender identity. Descriptors in \textcolor{gray}{gray} are only used in the open-ended tasks, since the OpinionsQA dataset for the closed-ended Survey Response task does not cover nonbinary gender identity. }
    \label{tab:desc_gender}
\end{table*}

\begin{table*}
\centering
\renewcommand{\arraystretch}{1.2} 
    \begin{tabular}{l|p{6.5cm}|p{6cm}}
    \toprule
        Placeholder & Open-ended tasks & Closed-ended task \\
    \midrule
        \textnormal{[R\_list1]} & White, Black, Asian, Middle-Eastern or Hispanic & White, Black, Asian or Hispanic \\
        \textnormal{[R\_list2]} & 'White', 'Black', 'Asian', 'Middle-Eastern' or 'Hispanic' & 'White', 'Black', 'Asian' or 'Hispanic' \\
        \textnormal{[G\_list1]} & a man, a woman or a nonbinary person & a man or a woman \\
        \textnormal{[G\_list2]} & 'male', 'female' or 'nonbinary' & 'male' or 'female' \\  
    \bottomrule
    \end{tabular}
    \caption{\textbf{Additional Descriptors.} We list all remaining descriptors that occur in the templates listed in Table \ref{tab:prompt_templates}.}
    \label{tab:desc_add}
\end{table*}

\section{Computational Details} \label{app:comp_details}
We use vLLM\footnote{\url{https://github.com/vllm-project/vllm}} version 0 (for the open-ended tasks) and version 1 (for OpinionsQA since it was faster) for LLM inference. We run the \texttt{Llama-3.1-8B-Instruct} and \texttt{OLMo-2-1124-7B-Instruct} models on a single Nvidia A100 (40GB) GPU with a total running time of approximately 9:50h (0:30h for self-descriptions, 0:20h for social media bios and 9:00h for OpinionsQA) per model. 
We run \texttt{Llama-3.3-70B-Instruct}, \texttt{OLMo-2-0325-32B-Instruct} and \texttt{gemma-3-27b-it} on two Nvidia H100 (80GB) GPUs in parallel. The total running time for generating all outputs with all three models is approximately 43:40h (5:40h for self-descriptions, 2h for social media bios, 36h for OpinionsQA).

For all models, we use the parameters recommended by the model creators. To generate answer option distributions for OpinionsQA, we extract temperature-scaled log probabilities, applying the temperature values recommended by the model creators.

\section{Preprocessing of LLM Responses} \label{sec:preprocessing}
For the open-text responses we apply a variety of measures to identify stereotypes and flattening effects, with the normative expectation that groups are not treated differently when generating self-descriptions or social media bios for them based on demographics alone. To accurately detect such patterns, we first remove all demographic markers, names, and pronouns from our prompts that have been repeated in the LLMs' answer. Using a regex, we replace them with the string \narrowfont{"[identity]"}, while ensuring that consecutive redactions are combined. For instance, we replace \narrowfont{"I am Ms. Garcia, a …"} with \narrowfont{"I am [identity], a …"}, when \narrowfont{"Ms."} and \narrowfont{"Garcia"} were both part of our prompt.

\subsection{Identification of Role Violations} \label{app:role_violations_ann}

We further filter the set of free-text responses to exclude cases in which the LLMs failed to maintain their assigned role. This includes instances where the model refused to answer (e.g., \narrowfont{"I cannot fulfill this request."}), answered as an AI assistant (e.g., \narrowfont{"Just a language model navigating the human world."}) or partially reverted to an AI identity after initially adopting the persona (e.g., \narrowfont{"I am a 32-year-old Black female. I identify as an AI..."}). 

To identify such role violations, we used \texttt{Qwen2.5-32B}~\cite{bai2023qwen}---a model not included in our main experiments to avoid self-preference bias in LLM annotations~\cite{wataoka2024self}---for automated annotation. We then validated this automatic classification by conducting a small annotation study, with 300 randomly sampled LLM responses where three human annotators who are fluent in English verified the labels. 
Both, the human annotators and \texttt{Qwen2.5-32B} received a similar instruction with the \textcolor{gray}{grey} part being only shown to the LLM:

\textit{You are an expert annotator tasked with classifying whether a self-description or social media bio was written by a language model or a human. Language models might describe themselves as AI, chat assistant, large language model, etc., while human self-description or social media bios are clearly describing a human being. If the self-description or social media bio contains a language model refusal, also classify it as "AI".\\
\textcolor{gray}{Only provide your annotation, no additional explanation! 
Use the following JSON format for your annotation: \{"author": <one of the following options: "AI" or "human", without any extra commentary whatsoever; just the annotation!>,\}. 
Now annotate the following self-description or social media bio:}}

The annotators were debriefed that the content they annotate does not contain any harmful or emotionally triggering content and that their annotations would be used for validation, to which they consented. Each annotator had to annotate 150 instances out of 300. The average Cohen’s $\kappa$ among the annotators was 0.925, indicating high inter-annotator agreement. When computing the accuracy of \texttt{Qwen2.5-32B} on the human annotated data, we found that it was 100\% accurate.  

\section{Additional Results}

\subsection{Statistics on Role Violations} \label{app:role_violations}
\begin{figure*}
    \centering
    \begin{subfigure}[b]{0.47\textwidth}
        \centering
        \includegraphics[width=\linewidth]{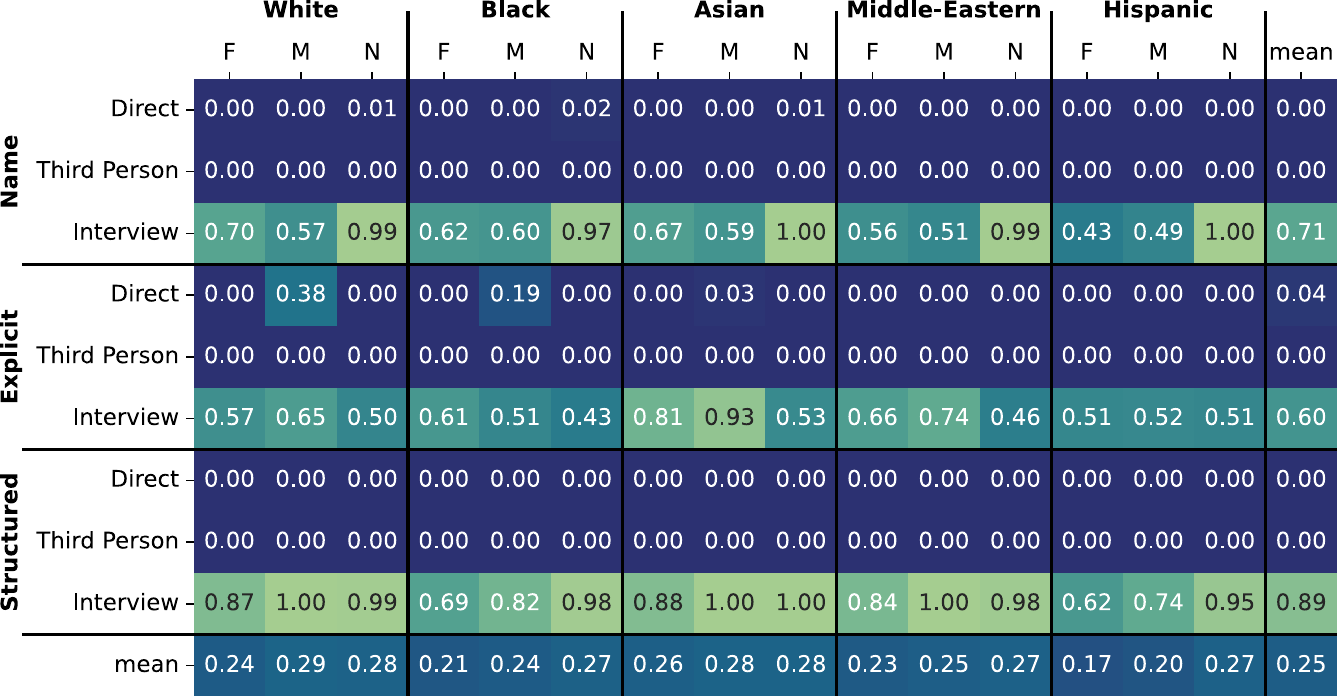}
        \caption{\texttt{gemma-3-27b-it} ($\downarrow$)}
        \label{}
    \end{subfigure}
    \hfill
    \begin{subfigure}[b]{0.52\textwidth}
        \centering
        \includegraphics[width=\linewidth]{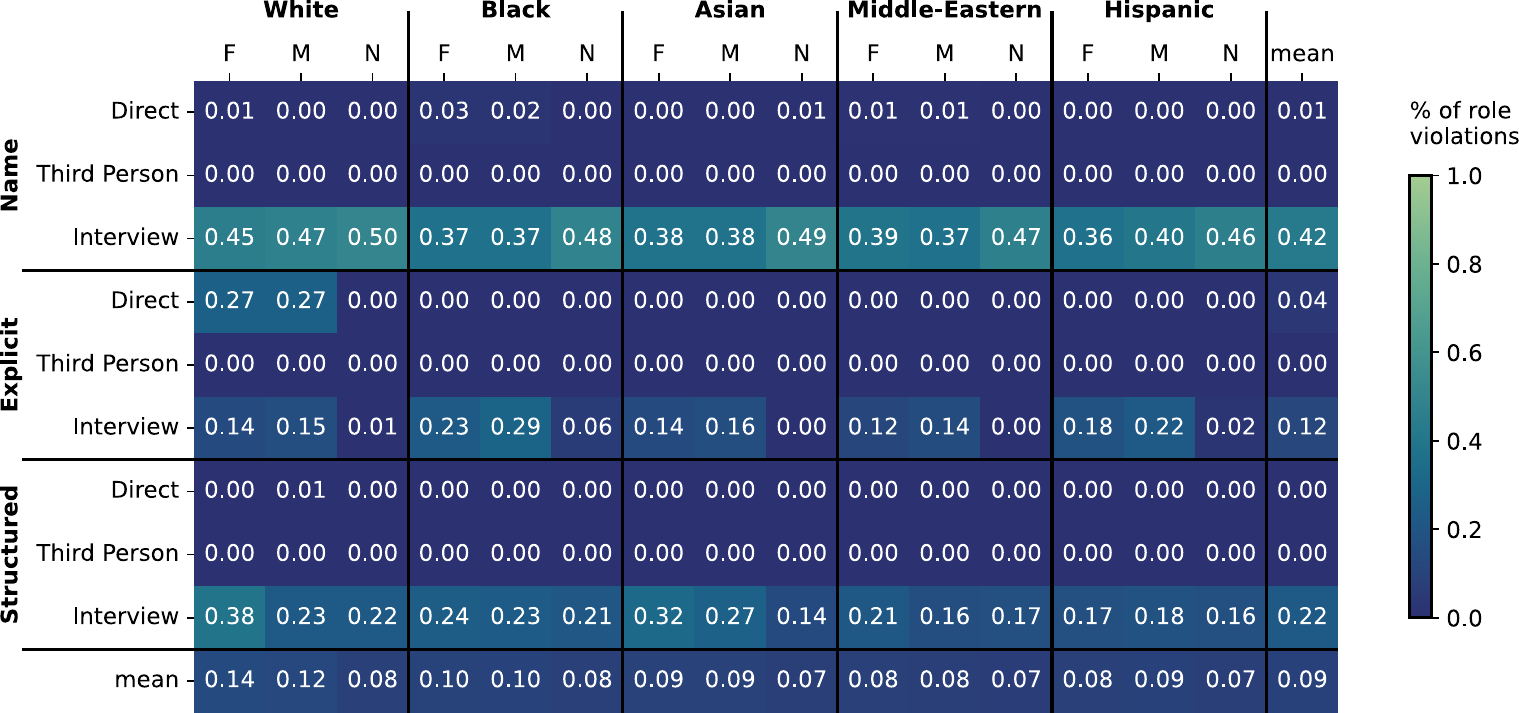}
        \caption{\texttt{Llama-3.1-8B-Instruct} ($\downarrow$)}
        \label{}
    \end{subfigure}
    \caption{ \textbf{Role violations in self-descriptions.} We report the rate of role violations across combinations of prompt types and demographic groups, focusing on the two models with the highest violation rates. Notably, Gemma-3-27b exhibits role violations reaching 100\% for some combinations, particularly for the \textit{interview} prompt format and \textit{nonbinary} personas.}
    \label{fig:role_vio}
\end{figure*}

We analyze the extent to which a model violates its assigned role, that is, it refuses to answer or answers with the identity of an AI assistant (see Appendix \ref{app:role_violations_ann} for details).
We find that Llama-3.3-70B, OLMo-2-32B-Instruct, and OLMo-2-7B result in no or very few role violations (up to 2.4\%). Results for Llama-3.1-8B and Gemma-3-27b are presented in Figure \ref{fig:role_vio}. We observe that the \textit{interview} format generally leads to a higher number of role violations, with Gemma-3-27b exhibiting notable difficulties in simulating personas for certain combinations of prompt types and demographic groups. We suspect that this is due to a conflict between the persona simulation instruction and the system prompt's directive to behave as an AI assistant.

\subsection{Social Media Bios}\label{app:bio}
\begin{figure*}[htbp]
    \centering
    \begin{subfigure}[b]{0.49\textwidth}
        \centering
        \includegraphics[width=\linewidth]{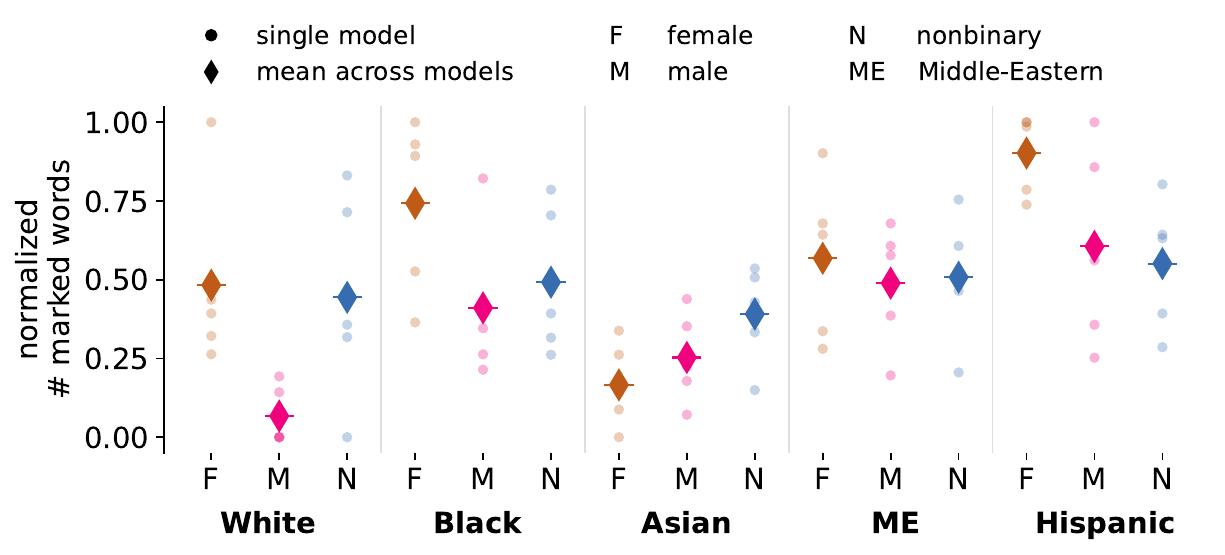}
        \caption{normalized marked word count ($\downarrow$)}
        \label{}
    \end{subfigure}
    \hfill
    \begin{subfigure}[b]{0.49\textwidth}
        \centering
        \includegraphics[width=\linewidth]{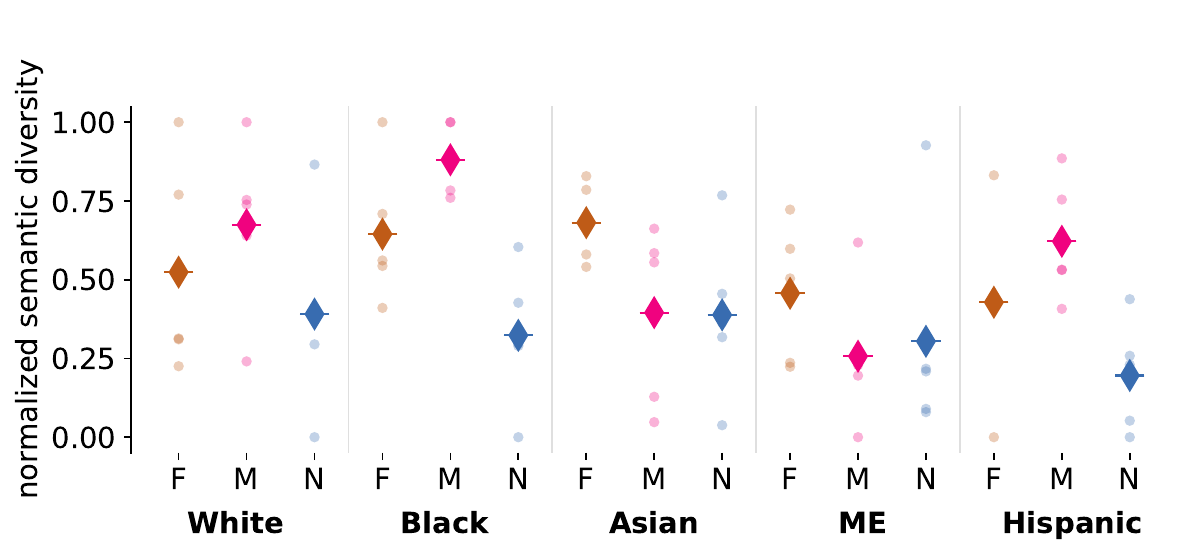}
        \caption{normalized semantic diversity ($\uparrow$)}
        \label{fig:sub_sdiv_bio}
    \end{subfigure}
    \caption{\textbf{Discrepancies between demographic groups in social media bios.} We show the (a) normalized number of marked words and (b) normalized semantic diversity of generated social media bios aggregated across prompt types for each demographic group. Min-max normalization was applied for each model separately to indicate the relative ranking of demographic groups. We observe that \textit{Middle-Eastern} and \textit{Hispanic} personas produce less favorable outputs than other race/ethnicity groups (i.e., higher marked word count and lower semantic diversity).}
    \label{fig:Sem_div_dem_Bio}
\end{figure*}

\begin{figure*}[htbp]
    \centering
    \begin{subfigure}[b]{0.49\textwidth}
        \centering
        \includegraphics[width=\linewidth]{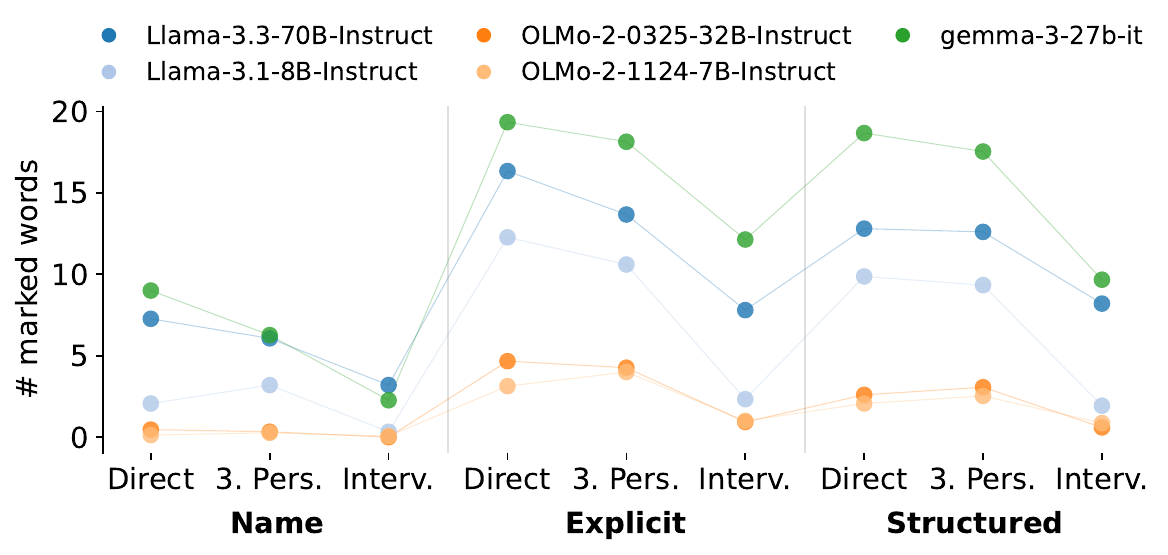}
        \caption{marked word count ($\downarrow$)}
        \label{fig:marked_words_model_bio}
    \end{subfigure}
    \hfill
    \begin{subfigure}[b]{0.49\textwidth}
        \centering
        \includegraphics[width=\linewidth]{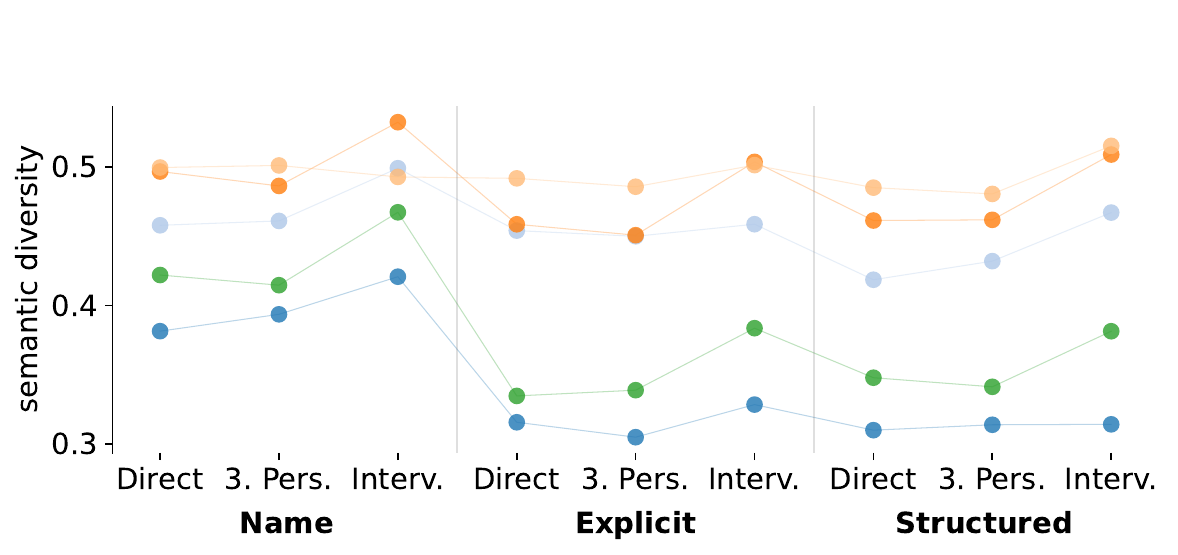}
        \caption{semantic diversity ($\uparrow$)}
        \label{fig:marked_words_norm_bio}
    \end{subfigure}
    \caption{ \textbf{Comparison of prompt types and models for social media bios.} We present the (a) number of marked words and (b) semantic diversity of generated social media bios for each prompt type and model. Values are aggregated across all demographic groups. We find that prompts using \textit{names} and the \textit{interview} format consistently lead to the lowest number of marked words and highest semantic diversity for all models. Further, we observe that Llama-3.3-70B and Gemma-3-27B lead to the worst results (i.e., high number of marked word and low semantic diversity), while OLMo-2-32B and OLMo-2-7B yield the best results (i.e., low number of marked word and high semantic diversity).
    }
    \label{fig:MW_Sem_div_Bio}
\end{figure*}

\begin{figure}
    \centering
    \includegraphics[width=\linewidth]{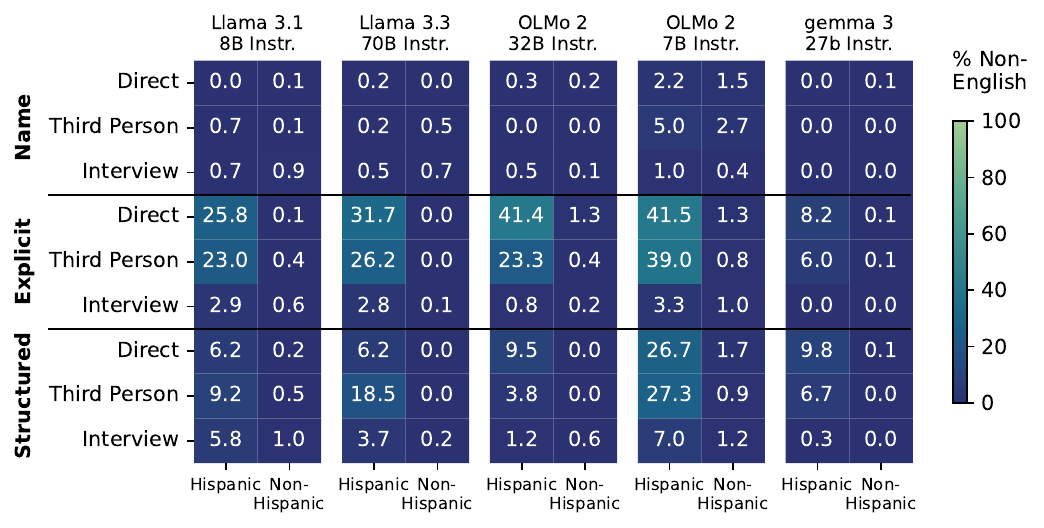}
    \caption{\textbf{Percentage of non-english social media bios ($\downarrow$)}. We report the percentage of non-English responses generated for \textit{Hispanic} personas, who receive the highest proportion of such responses compared to other race/ethnicitiy groups. We observe that \textit{explicit} and \textit{structured} demographic priming leads to higher rates of non-english responses than using \textit{names}, which yields rates close to 0. Notably, the \textit{interview} format appears to mitigate language switching with respect to \textit{explicit} and \textit{structured} demographic priming.}
    \label{fig:lang_bio}
\end{figure}

Complementary evaluation results for social media bios are presented in Figures \ref{fig:Sem_div_dem_Bio}, \ref{fig:MW_Sem_div_Bio}, and \ref{fig:lang_bio}. We find that the results overall align with those reported for self-descriptions.

\subsection{Classification Accuracy} \label{app:marked_words}
\begin{figure*}[htbp]
    \centering
    \begin{subfigure}[b]{0.49\textwidth}
        \centering
        \includegraphics[width=\linewidth]{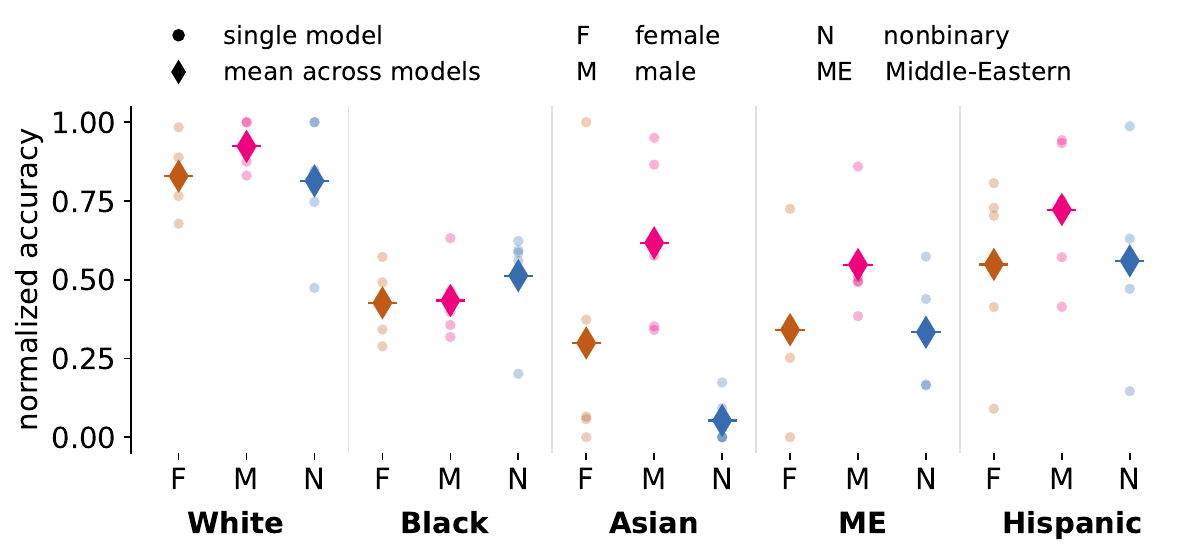}
        \caption{accuracy in self-descriptions}
    \end{subfigure}
    \hfill
    \begin{subfigure}[b]{0.49\textwidth}
        \centering
        \includegraphics[width=\linewidth]{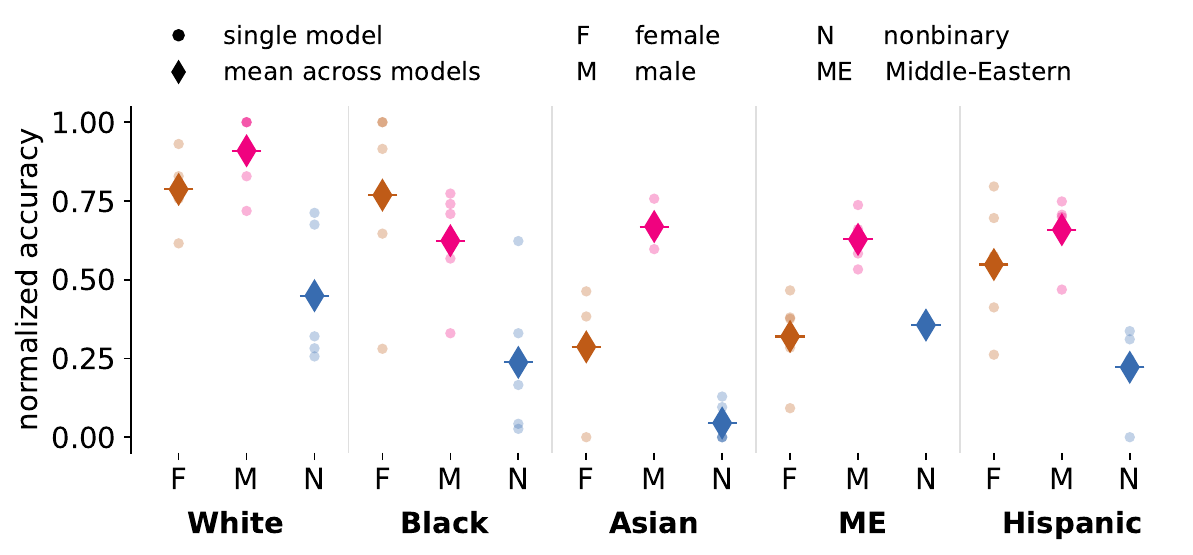}
        \caption{accuracy in social media bios}
    \end{subfigure}
    \caption{\textbf{Normalized classification accuracy across demographic groups ($\downarrow$).} We present the classification accuracy across demographic groups. Values are averaged over all prompt types with lower values indicating lower distinguishability between demographic groups. 
    In contrast to the other open-text evaluation measures, we find that personas with \textit{male} gender and \textit{White} race/ethnicity are easiest to classify in both tasks, while \textit{nonbinary} personas are hardest to detect. However, we note that the absolute accuracy for all groups is generally high ($>0.9$), indicating that responses of all groups are relatively easy to classify.}
    \label{fig:acc_all_tasks_dem}
\end{figure*}

\begin{figure*}[htbp]
    \centering
    \begin{subfigure}[b]{0.49\textwidth}
        \centering
        \includegraphics[width=\linewidth]{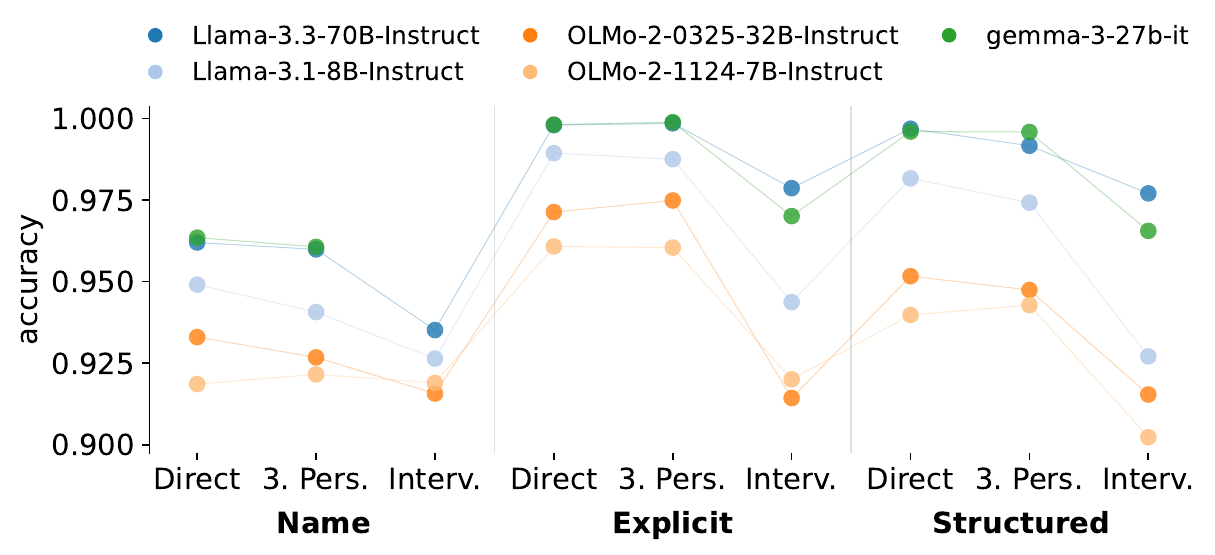}
        \caption{accuracy in self-descriptions}
        \label{fig:acc_model_SD}
    \end{subfigure}
    \hfill
    \begin{subfigure}[b]{0.49\textwidth}
        \centering
        \includegraphics[width=\linewidth]{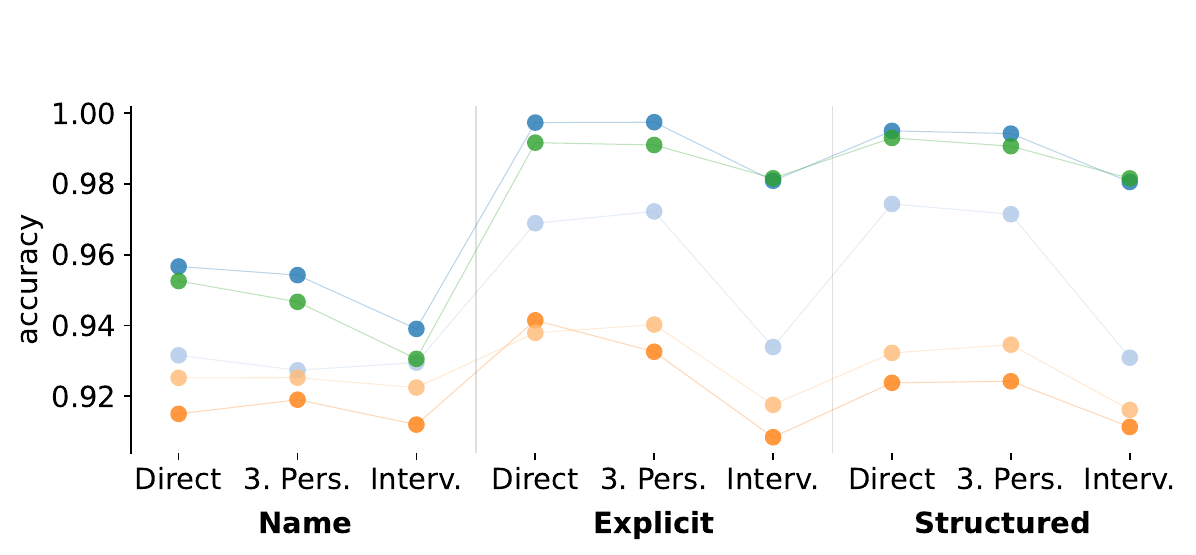}
        \caption{accuracy in social media bios}
        \label{fig:acc_model_bio}
    \end{subfigure}
    \caption{ \textbf{Classification accuracy across prompt types and models ($\downarrow$).} We present the classification accuracy across different prompt types. Values are averaged over all demographic groups with lower values indicating lower distinguishability between demographic groups. 
     We observe a consistent pattern: prompts using \textit{names} and the \textit{interview} format result in the lowest (i.e., best) accuracy on average for all models across both tasks. We further find that responses from Gemma-3-27b and Llama-3.3-70B consistently lead to higher accuracy on average across both tasks, while OLMo-2-32B and OLMo-2-7B yield the lowest.}
    \label{fig:acc_all_tasks}
\end{figure*}

We show the classification accuracy disaggregated by demographic group and prompt type in Figures \ref{fig:acc_all_tasks_dem} and \ref{fig:acc_all_tasks}. We note that the overall high accuracy values ($> 0.9$) suggest that predicting demographics from simulated self-descriptions and social media bios is an easy task for LLMs, aligning with \citet{cheng2023marked}. This may also indicate that responses remain specific to each demographic group rather than being homogenized.

\subsection{Disparities Between Demographic Groups} \label{app:var_dem}
We investigate how quantitative measures vary across demographic groups by computing the standard deviation of each measure between groups, where lower values indicate reduced disparities. To analyze the impact of prompt types on these disparities, we conduct ordinary least squares (OLS) regression analyses using the standard deviation as the dependent variable (results shown in Table~\ref{tab:std}). Our goal is to assess whether the observed benefits of prompting with \textit{names} and using the \textit{interview} format translate into to more equitable outcomes across demographic groups.

We find that prompting with \textit{names} and the \textit{interview} format with OLMo-2-7B and OLMo-2-32B leads to a significant reduction in group disparities across all open-text measures -- except for accuracy, where no significant change is observed. For the other models and the the closed-ended task, results across measures are more mixed, suggesting that while these prompt strategies seem to improve the overall representation of demographic personas, their positive effects may not always be uniform, benefiting some groups more than others. 

\begin{table*}[htb]
  \centering
  \begin{subtable}[b]{0.48\textwidth}
    \centering
    \scriptsize
    \setlength{\tabcolsep}{2pt}  
    \begin{tabular}{l d{-.3} d{-.3} d{-.3} d{-.3} d{-.3}}
        \toprule
        & \multicolumn{1}{c}{Llama-70B} & \multicolumn{1}{c}{Llama-8B} & \multicolumn{1}{c}{OLMo-32B} & \multicolumn{1}{c}{OLMo-7B} & \multicolumn{1}{c}{Gemma-27b} \\
        \midrule
        \textcolor{okabe_orange}{Name} & -.014 & -.002 & -.007* & -.010* & -.000 \\
        \textcolor{okabe_orange}{Struct.} & -.008 & -.004 & -.001 & -.004 & -.002 \\
        \textcolor{okabe_red}{Interview} & .016 & .005 & -.007* & -.011* & .008* \\
        \textcolor{okabe_red}{3. Person} & .012 & .003 & .000 & .000 & .003 \\
        \midrule
        {Self-Descr.} & -.018* & .009* & .010* & .013* & -.007* \\
        Phrasing v2 & .003 & .001 & .001 & .003 & .004 \\
        \midrule
        Intercept & .066* & .036* & .022* & .033* & .039* \\
        \bottomrule
    \end{tabular} 
    \caption{semantic diversity}
    \vspace{1em}
    \label{tab:std_sd}
    \end{subtable}
    \hfill
    \begin{subtable}[b]{0.48\textwidth}
    \centering
    \scriptsize
    \setlength{\tabcolsep}{2pt}  
    \begin{tabular}{l d{-.3} d{-.3} d{-.3} d{-.3} d{-.3}}
        \toprule
        & \multicolumn{1}{c}{Llama-70B} & \multicolumn{1}{c}{Llama-8B} & \multicolumn{1}{c}{OLMo-32B} & \multicolumn{1}{c}{OLMo-7B} & \multicolumn{1}{c}{Gemma-27b} \\
        \midrule
        \textcolor{okabe_orange}{Name} & -.127* & -.061* & -.157* & -.076* & -.092* \\
        \textcolor{okabe_orange}{Struct.} & -.103* & -.042* & -.150* & -.037 & -.049 \\
        \textcolor{okabe_red}{Interview} & -.040 & -.008 & -.102* & -.068* & -.069* \\
        \textcolor{okabe_red}{3. Person} & .049 & -.005 & -.014 & -.017 & .002 \\
        \midrule
        {Self-Descr.} & -.010 & -.022* & .055 & -.059* & .056* \\
        Phrasing v2 & .026 & .011 & .032 & .022 & .001 \\
        \midrule
        Intercept & .119* & .074* & .164* & .134* & .089* \\
        \bottomrule
    \end{tabular}
    \caption{share of non-English responses}
    \vspace{1em}
    \label{tab:std_lang}
  \end{subtable}
    \vfill
  \begin{subtable}[b]{0.48\textwidth}
    \centering
    \scriptsize
    \setlength{\tabcolsep}{2pt}  
    \begin{tabular}{l d{-.3} d{-.3} d{-.3} d{-.3} d{-.3}}
        \toprule
        & \multicolumn{1}{c}{Llama-70B} & \multicolumn{1}{c}{Llama-8B} & \multicolumn{1}{c}{OLMo-32B} & \multicolumn{1}{c}{OLMo-7B} & \multicolumn{1}{c}{Gemma-27b} \\
        \midrule
        \textcolor{okabe_orange}{Name} & .008* & .003 & -.002 & -.003 & .012* \\
        \textcolor{okabe_orange}{Struct.} & .000 & .001 & -.001 & .000 & .004 \\
        \textcolor{okabe_red}{Interview} & .003 & .005* & -.000 & .001 & .020* \\
        \textcolor{okabe_red}{3. Person} & -.001 & .000 & .002 & .000 & -.000 \\
        \midrule
        {Self-Descr.} & -.000 & .002 & .002 & .001 & .008 \\
        Phrasing v2 & -.000 & .002 & -.001 & .002 & .001 \\
        \midrule
        Intercept & .006* & .011* & .016* & .015* & -.000 \\
        \bottomrule
    \end{tabular} 
    \caption{accuracy}
    \label{tab:std_acc}
    \end{subtable}
    \hfill
    \begin{subtable}[b]{0.48\textwidth}
    \centering
    \scriptsize
    \setlength{\tabcolsep}{2pt}  
    \begin{tabular}{l d{-.3} d{-.3} d{-.3} d{-.3} d{-.3}}
        \toprule
        & \multicolumn{1}{c}{Llama-70B} & \multicolumn{1}{c}{Llama-8B} & \multicolumn{1}{c}{OLMo-32B} & \multicolumn{1}{c}{OLMo-7B} & \multicolumn{1}{c}{Gemma-27b} \\
        \midrule
        \textcolor{okabe_orange}{Name} & -.000 & .000 & .001 & .001 & .001 \\
        \textcolor{okabe_orange}{Struct.} & .000 & -.002* & .000 & .001 & -.001 \\
        \textcolor{okabe_red}{Interview} & .002* & -.001* & .002* & -.001 & .001 \\
        \textcolor{okabe_red}{3. Person} & .003* & -.000 & -.001 & .000 & -.001 \\
        \midrule
        Phrasing v2 & .001 & .000 & -.001 & -.001 & -.001 \\
        \midrule
        Intercept & .016* & .016* & .016* & .013* & .018* \\
        \bottomrule
    \end{tabular}
    \caption{opinion distance}
    \label{tab:std_dist}
  \end{subtable}
  \caption{{\textbf{Regressions on the standard deviation of all quantitative measures.} We conduct OLS regression analyses per LLM, using the standard deviation of each quantitative measure as a dependent variable and report the regression coefficients. Lower standard deviation~($\downarrow$) indicates reduced disparities between demographic groups. The independent variables include: \textcolor{okabe_orange}{demographic priming} (reference: explicit), \textcolor{okabe_red}{role adoption} (reference: direct), prompt phrasing (reference: v1), and, for the open-text measures, {task} (reference: Bio). *~$p < 0.05$.}}
  \label{tab:std}
\end{table*}

\subsection{Alignment with OpinionsQA} \label{app:OpinionQA}
\begin{figure*}[htbp]
    \centering
    \begin{subfigure}[b]{0.49\textwidth}
        \centering
        \includegraphics[width=\linewidth]{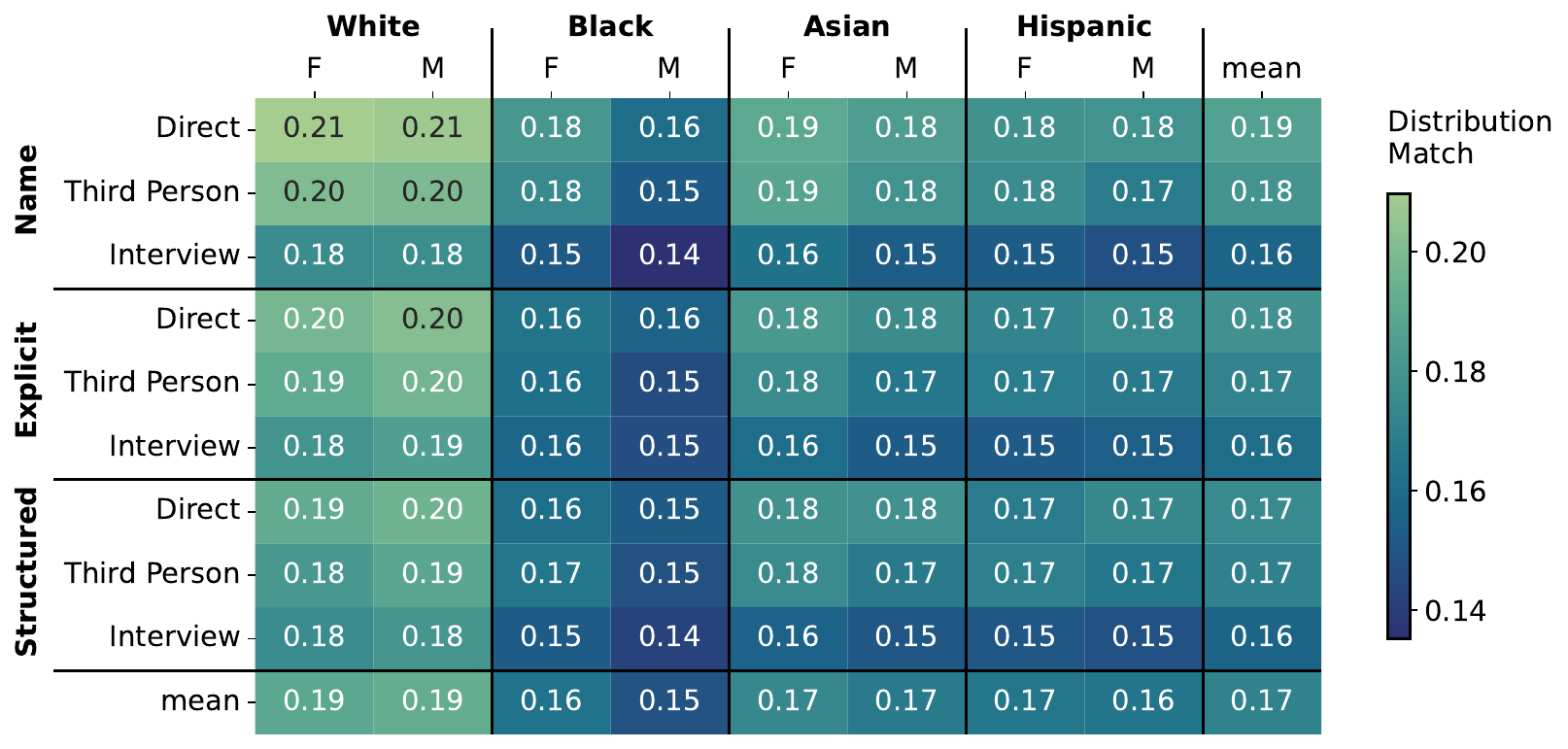}
        \caption{\texttt{Llama-3.1-8B-Instruct}}
        \label{}
    \end{subfigure}
    \hfill
    \begin{subfigure}[b]{0.49\textwidth}
        \centering
        \includegraphics[width=\linewidth]{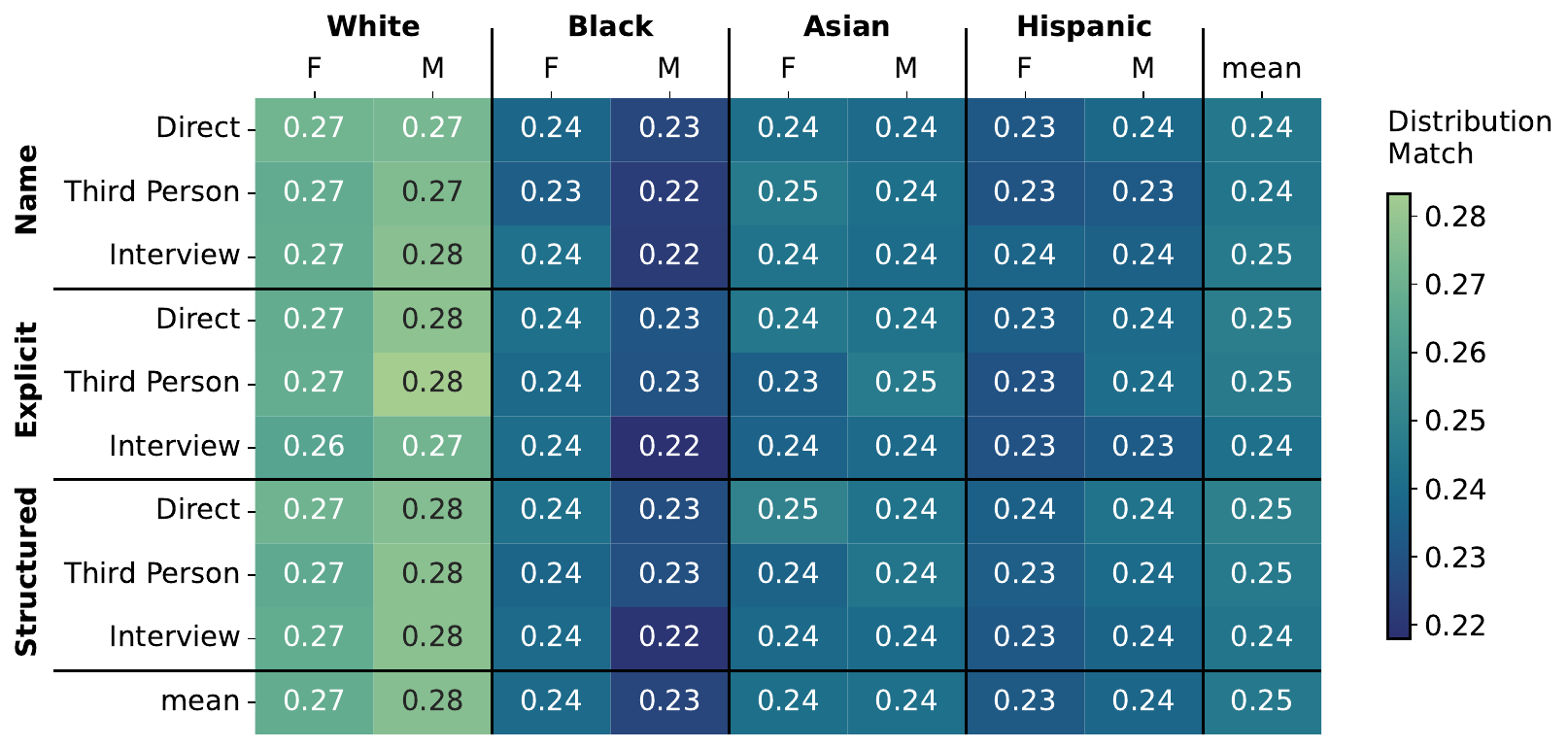}
        \caption{\texttt{Llama-3.3-70B-Instruct}}
        \label{}
    \end{subfigure}
    
    \begin{subfigure}[b]{0.49\textwidth}
        \centering
        \includegraphics[width=\linewidth]{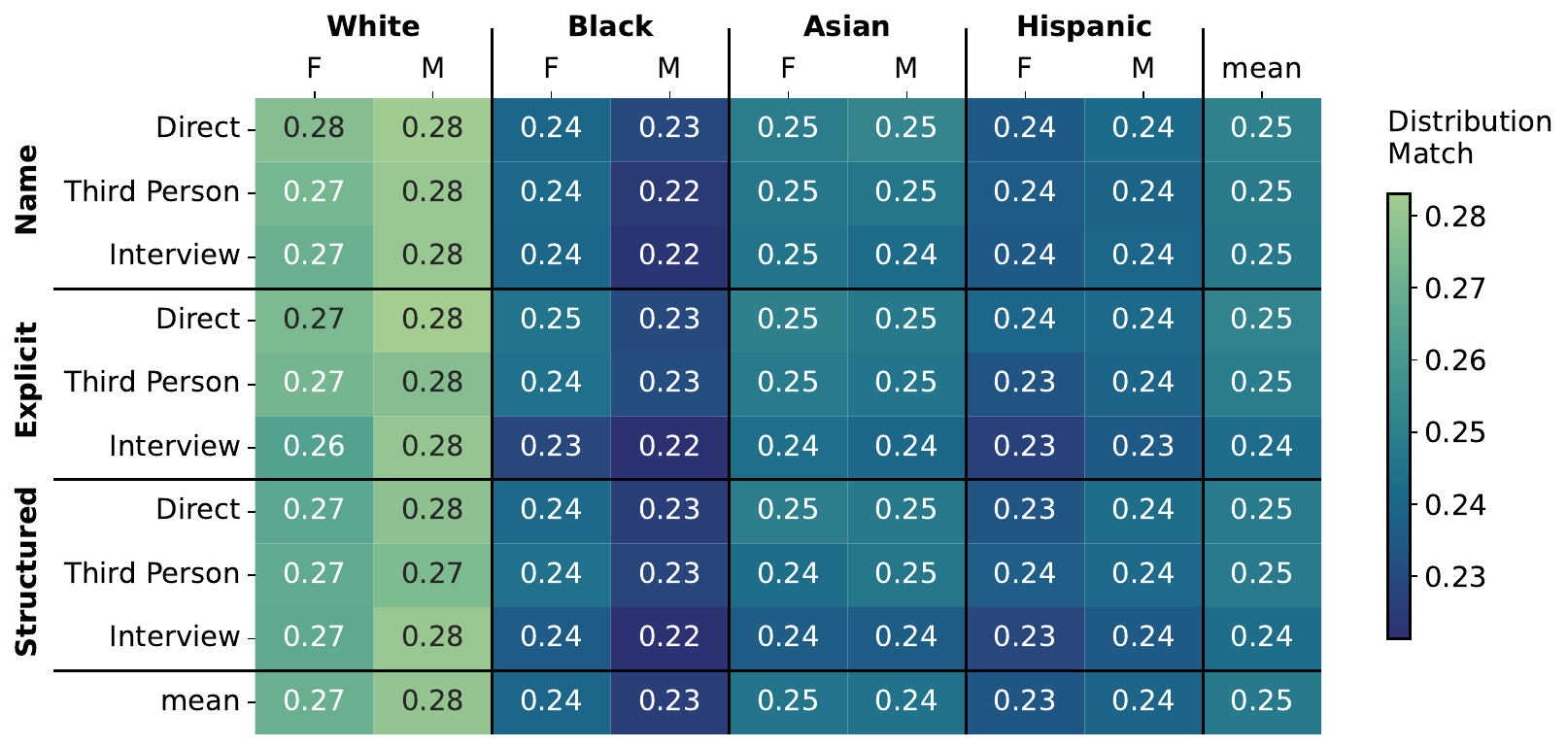}
        \caption{\texttt{gemma-3-27b-it}}
        \label{}
    \end{subfigure}
    \hfill
    \begin{subfigure}[b]{0.49\textwidth}
        \centering
        \includegraphics[width=\linewidth]{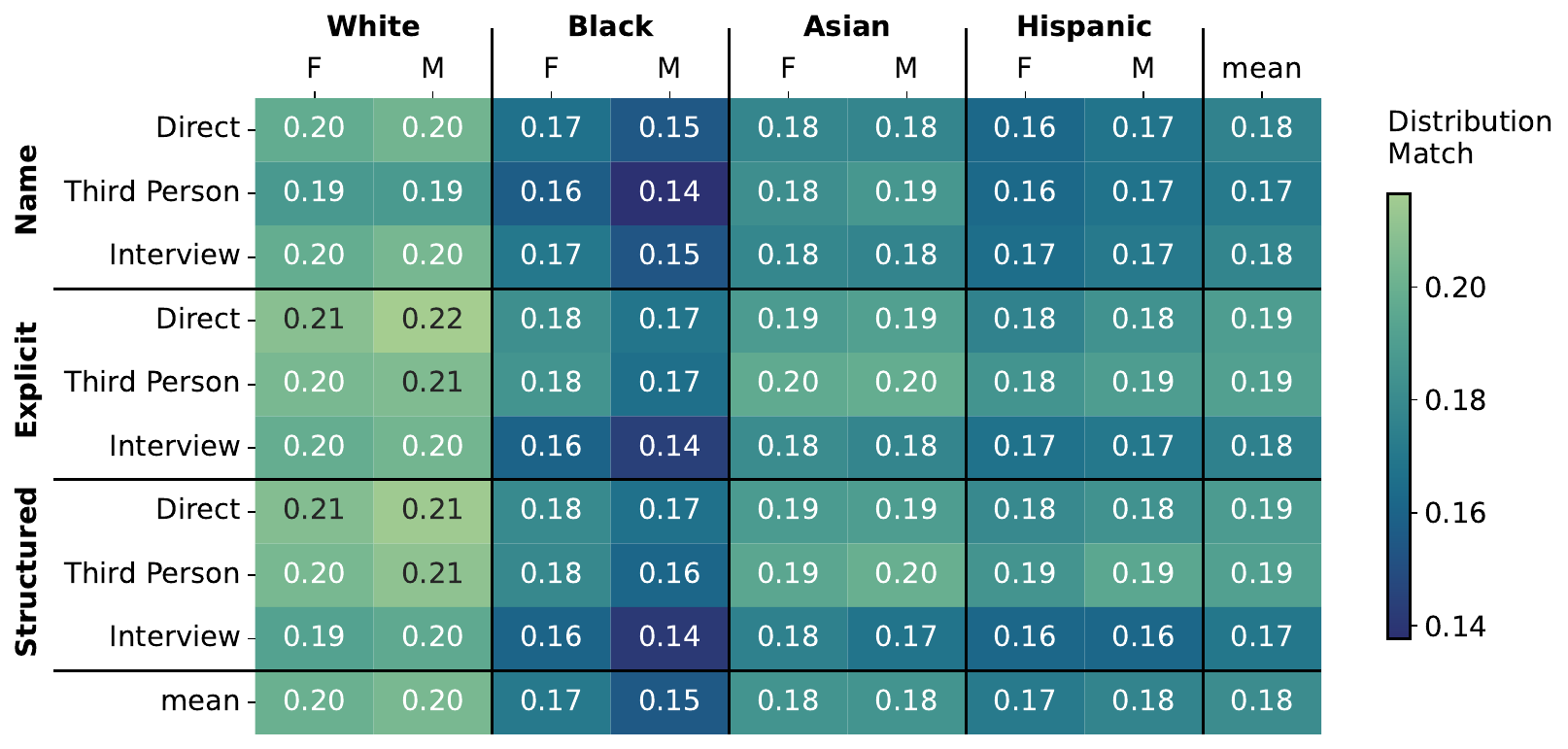}
        \caption{\texttt{OLMo-2-0325-32B-Instruct}}
        \label{}
    \end{subfigure}
    \caption{ \textbf{Opinion distance on OpinionsQA for all models ($\downarrow$).} We show the opinion distance (as measured by Wasserstein distance) across demographic groups and prompt types (lower is better). We observe that all models generally align better with the answer distribution of people with race/ethnicity \textit{Black}, while the highest opinion distance can be observed for people with race/ethnicity \textit{White}. We further observe that the average opinion distance for Llama-3.3-70B and Gemma-3-27b is higher (i.e., worse) or equal to the random baseline (0.25 $\pm$ 0.002) for most prompt types and demographic groups.}
    \label{fig:QA_dist}
\end{figure*}

We show the opinion distance (i.e., Wasserstein distance) on OpinionsQA for all remaining models in Figure \ref{fig:QA_dist}.

\begin{figure*}[htbp]
    \centering
    \begin{subfigure}[b]{0.49\textwidth}
        \centering
        \includegraphics[width=\linewidth]{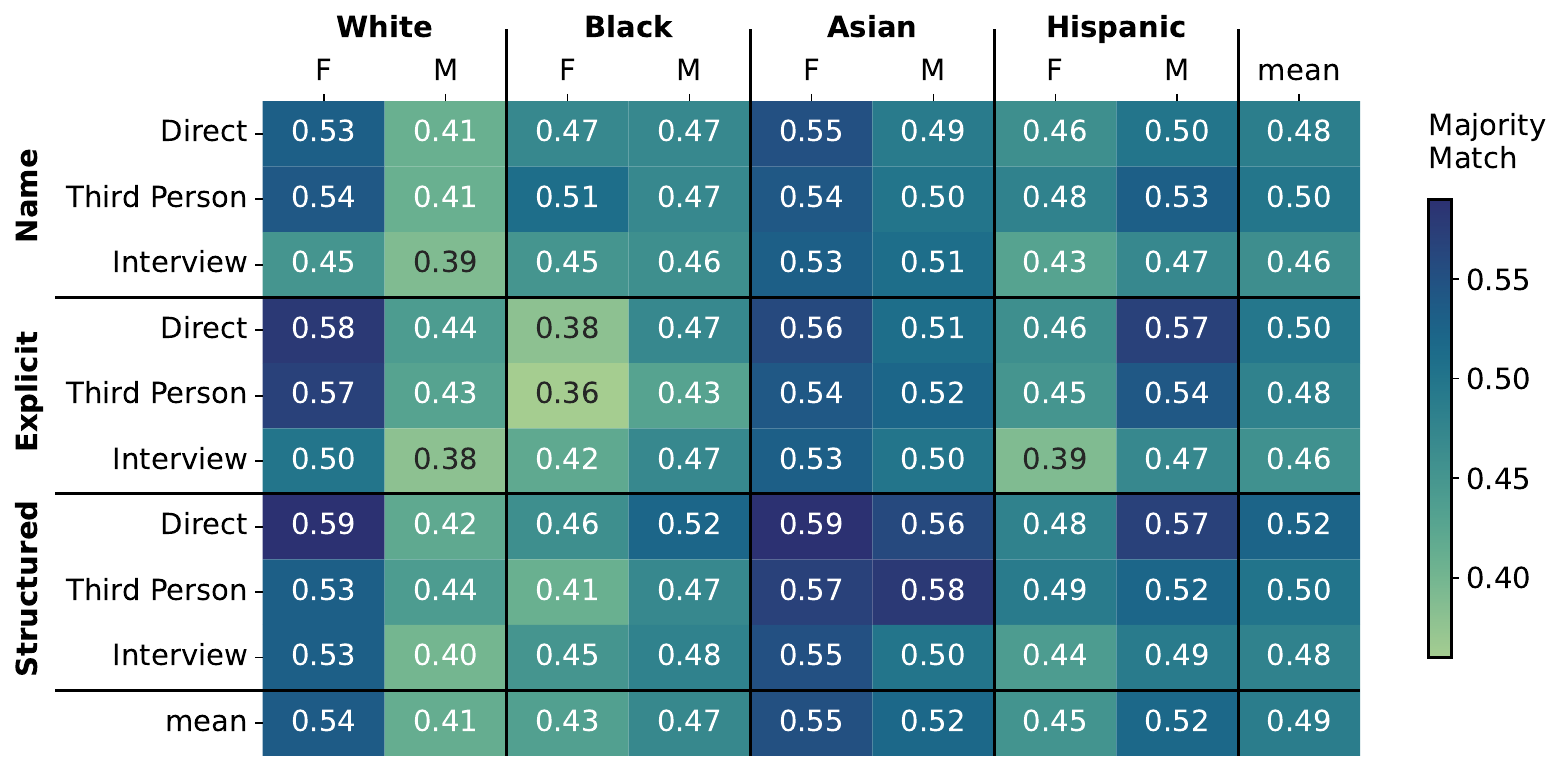}
        \caption{\texttt{Llama-3.1-8B-Instruct}}
        \label{}
    \end{subfigure}
    \hfill
    \begin{subfigure}[b]{0.49\textwidth}
        \centering
        \includegraphics[width=\linewidth]{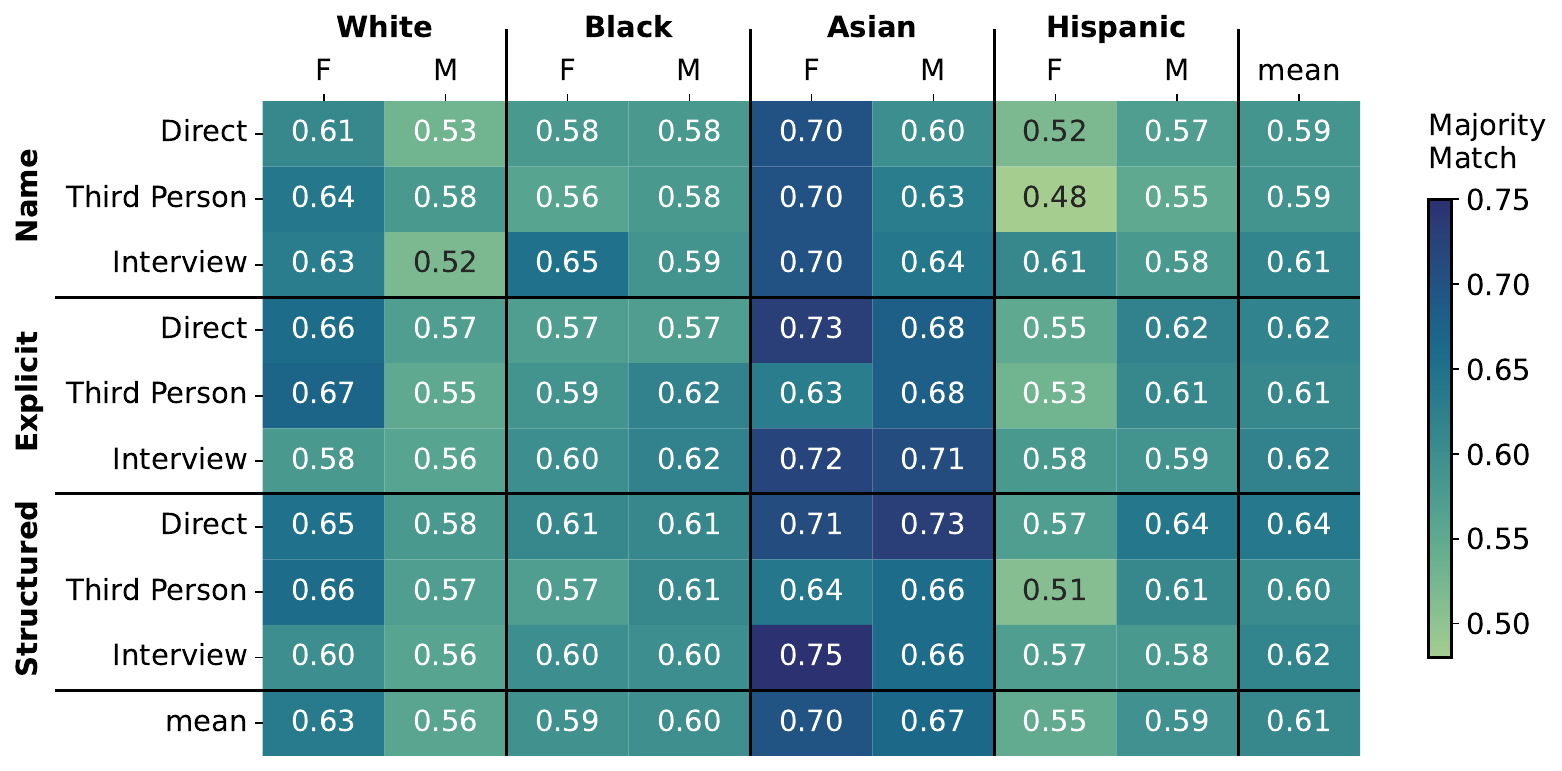}
        \caption{\texttt{Llama-3.3-70B-Instruct}}
        \label{}
    \end{subfigure}
    
    \begin{subfigure}[b]{0.49\textwidth}
        \centering
        \includegraphics[width=\linewidth]{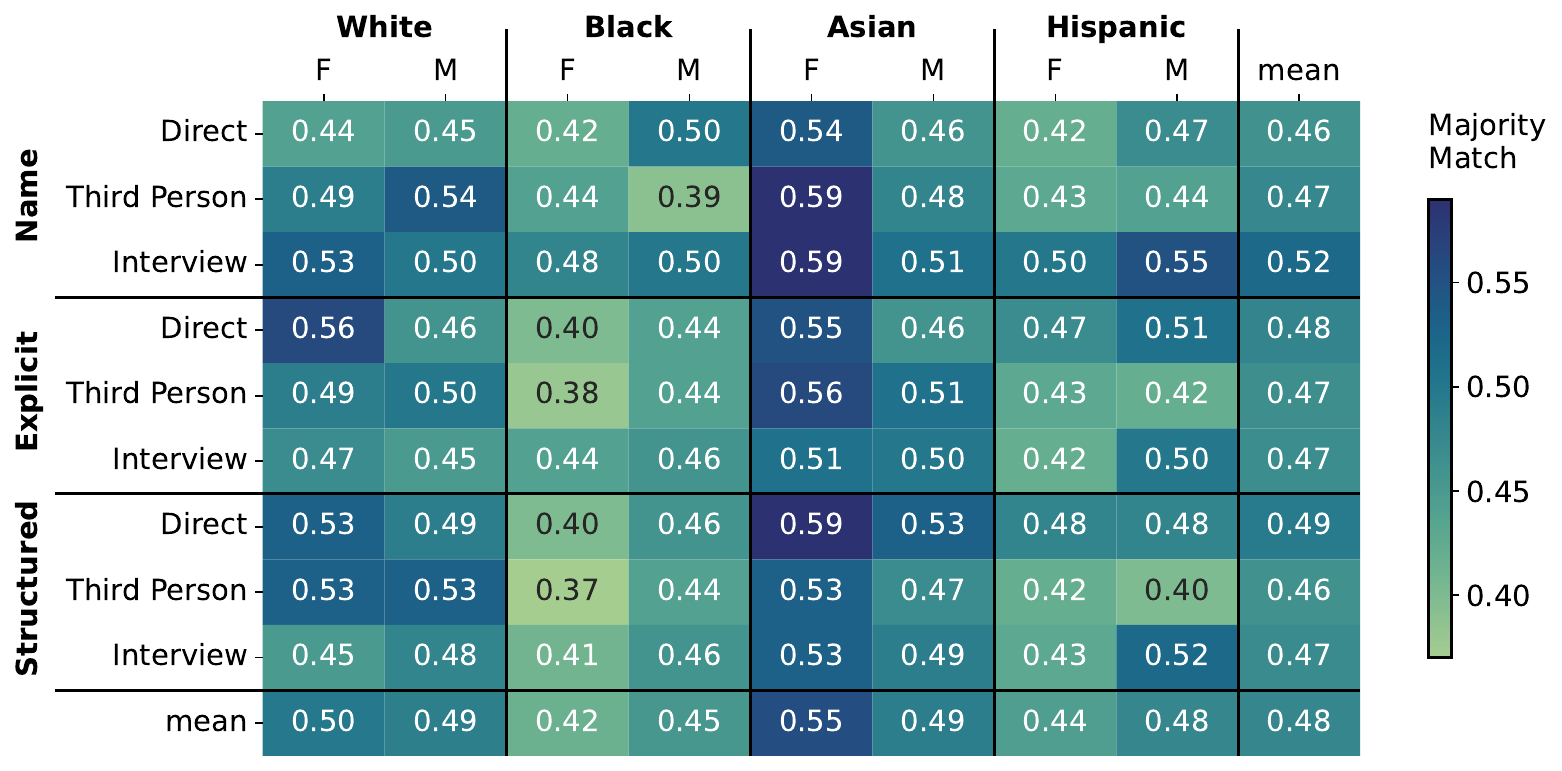}
        \caption{\texttt{gemma-3-27b-it}}
        \label{}
    \end{subfigure}
    \hfill
    \begin{subfigure}[b]{0.49\textwidth}
        \centering
        \includegraphics[width=\linewidth]{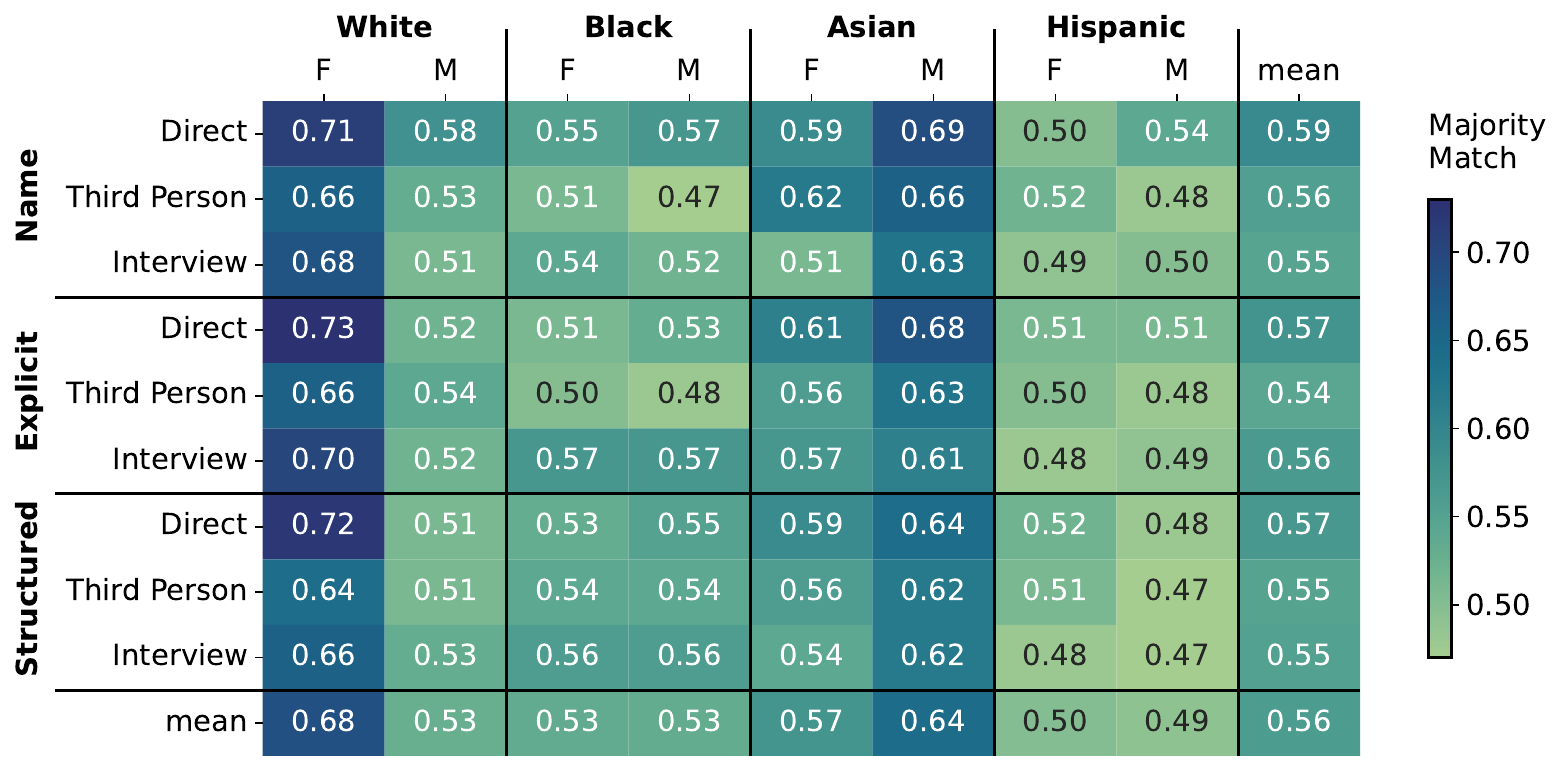}
        \caption{\texttt{OLMo-2-0325-32B-Instruct}}
        \label{}
    \end{subfigure}
    \begin{subfigure}[b]{0.49\textwidth}
        \centering
        \includegraphics[width=\linewidth]{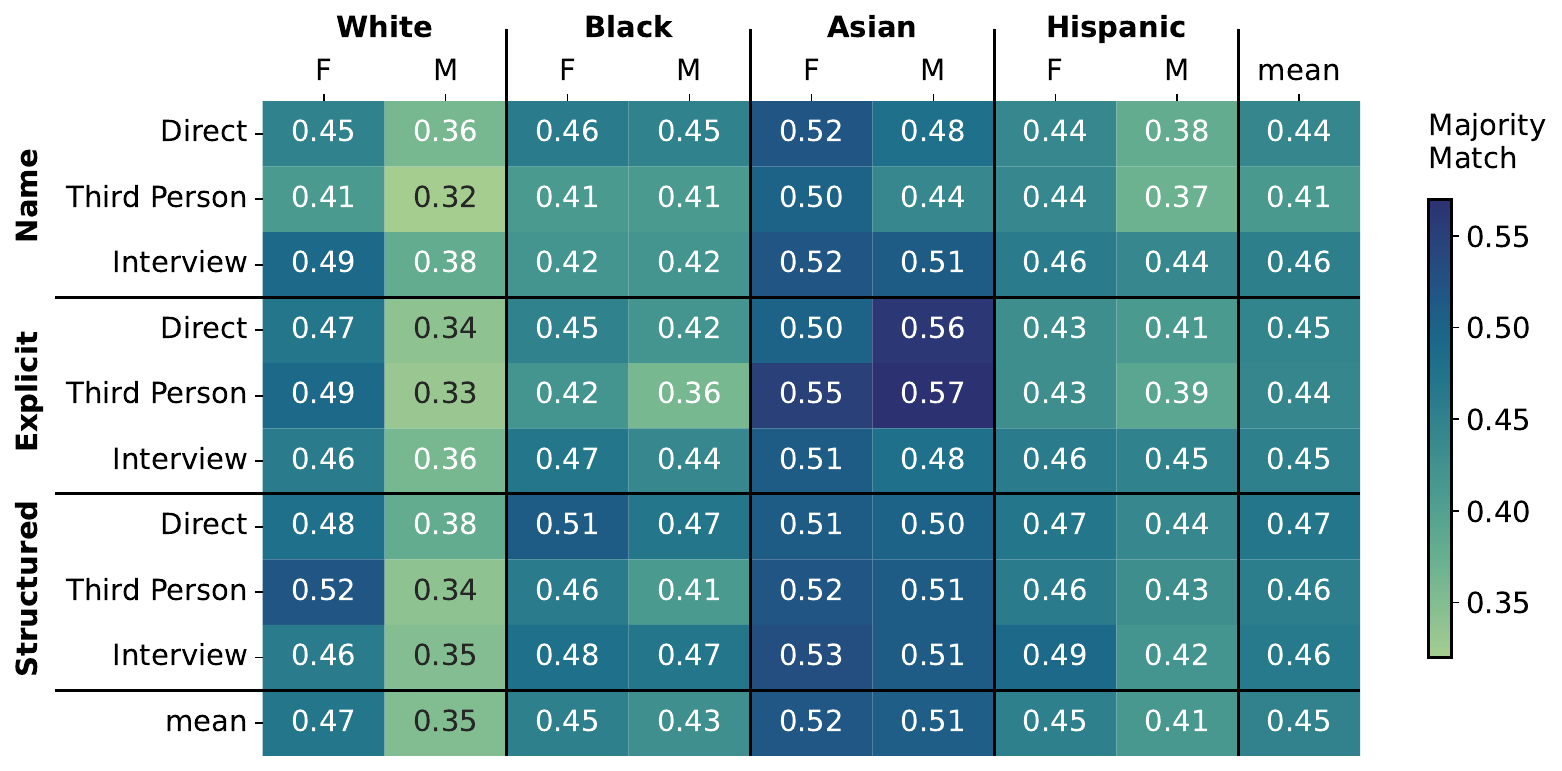}
        \caption{\texttt{OLMo-2-1124-7B-Instruct}}
        \label{fig:opinionqa_llama70}
    \end{subfigure}
    \caption{ \textbf{Majority match in OpinionsQA for all models ($\uparrow$).} We show the average majority match across demographic groups and prompt types (higher is better). We observe that most models generally align better with the responses of people with race/ethicity \textit{Asian} and with \textit{White females}. We further observe that the average majority match for all models is significantly higher (i.e., better) than the random baseline (0.24 $\pm$ 0.026) for most prompt types and demographic groups.}
    \label{fig:QA_maj}
\end{figure*}

Additionally, we also report \textbf{majority option match}, where we compute the option with the highest probability and compute the match of this single option between models and humans. Unlike opinion distance, here higher values indicate better alignment with the human responses.\footnote{Note that the majority option match is similar to, but not the exact same as, average accuracy over the 100 questions, since not all questions have the same answer options.} 
We show the majority match on OpinionsQA for all models in Figure \ref{fig:QA_maj}.

\subsection{Analysis of Stereotype Categories} \label{app:ster_cat}
\begin{figure*}
    \centering
    \begin{subfigure}[b]{0.49\textwidth}
    \centering
        \includegraphics[width=\linewidth]{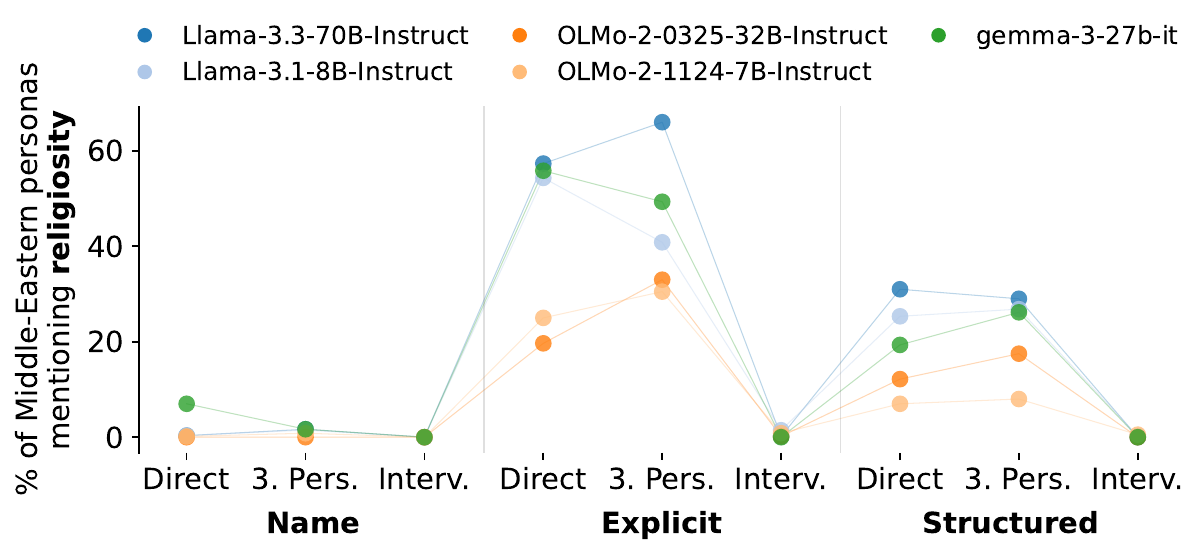}  
        \caption{conflation of identity with religion}
    \end{subfigure}
    \hfill
    \begin{subfigure}[b]{0.49\textwidth}
    \centering
        \includegraphics[width=\linewidth]{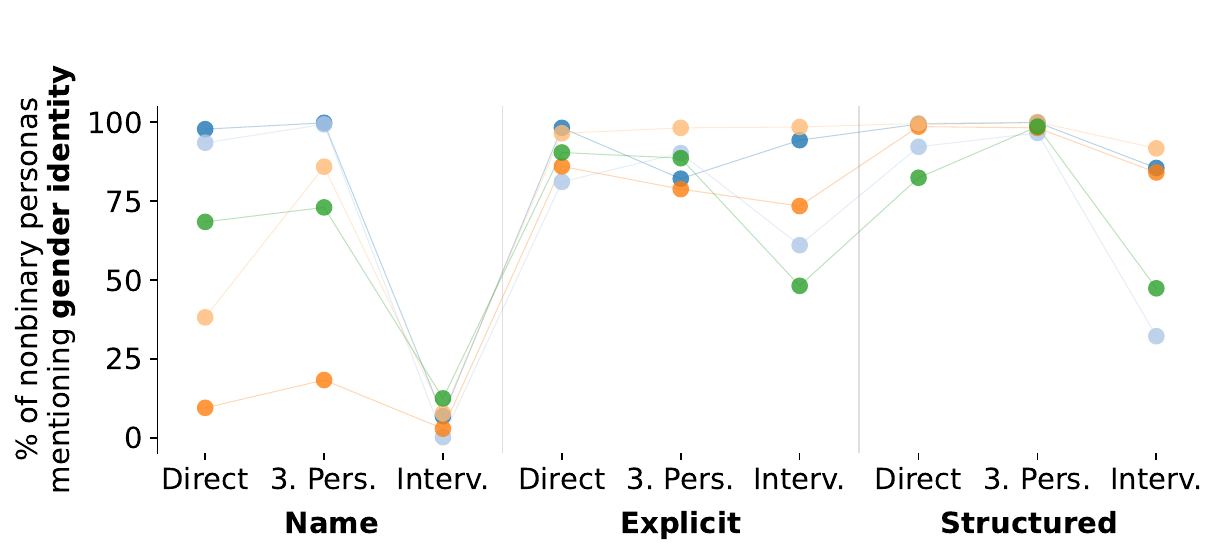}  
        \caption{disproportionate focus on gender identity}
    \end{subfigure}

    \caption{\textbf{Share of stereotyped persona descriptions across prompt types ($\downarrow$).} We show the share of self-descriptions by (a) \textit{Middle-Eastern} personas that contains words linked to religiosity (e.g., faith, muslim) and (b) \textit{nonbinary} personas containing terms focusing on gender identity (e.g., gender, identity). We find that both using \textit{names} and the \textit{interview} format reduces the share of self-descriptions reflecting these stereotype categories.} 
    \label{fig:genderidentity}
\end{figure*}

Figure \ref{fig:genderidentity} shows the share of self-descriptions by personas that are affected by the remaining stereotype categories as discussed in section \ref{sec:analysis_of_mw}. We observe that using \textit{names} and the \textit{interview} format consistently lowers the share of stereotyped persona descriptions across models. We test for significance with linear regressions per model and stereotype category, using role adoption and demographic priming as independent variables (reference: \textit{direct} and \textit{explicit}), and the share of stereotyped persona descriptions as dependent variable. Coefficients for \textit{interview} and \textit{name} are significantly negative (p < 0.001) across all models and categories, confirming that these strategies significantly reduce the expression of the specified problematic stereotypes in self-descriptions.

\section{Robustness Checks}
\subsection{Regression Analysis} \label{app:regression}
To assess whether differences across demographic groups and prompt types are statistically significant, we perform ordinary least squares (OLS) regression analyses using each evaluation measure as a dependent variable. We show the regression coefficients in tables \ref{tab:reg_mw}, \ref{tab:reg_sd}, \ref{tab:reg_lang}, \ref{tab:reg_acc} and \ref{tab:reg_dist}.

\paragraph{Influence of Prompt Types.}
For all open-text measures (i.e., marked word count, semantic diversity, share of non-english responses, and accuracy), we observe that demographic priming using \textit{names} leads to statistically significant improvements across \textbf{all} models ($p < 0.001$). Role adoption through the \textit{interview} format also yields significant improvements in marked word count and accuracy for all models ($p < 0.001$), and for 3 out of 5 models in the case of semantic diversity ($p < 0.001$) and share of non-English responses ($p < 0.01$).

With respect to the closed-ended task, we find that the coefficients for role adoption with the \textit{interview} format are negative and statistically significant ($p < 0.001$) for 4 out of 5 models, indicating that the \textit{interview} format significantly improves alignment. Further, \textit{name}-based prompting that reduced stereotypes in open-ended tasks does not significantly harm alignment in 4 out of 5 models.

\paragraph{Influence of Prompt Phrasing.}
To verify that our findings stem from the introduced prompt dimensions rather than specific wording, we include two alternative phrasings for each prompt template (cf. Table \ref{tab:prompt_templates}) and assess the impact of prompt phrasing on the results. We find that regression coefficients for prompt phrasing are significant in only 1 out of 5 models for the marked word count and the share of non-English responses (cf. Tables \ref{tab:reg_mw}, \ref{tab:reg_lang}), in 2 out of 5 models for opinion distance (cf. Table \ref{tab:reg_dist}), and in 3 out of 5 models for semantic diversity and accuracy (cf. Tables \ref{tab:reg_sd}, \ref{tab:reg_acc}). Across all measures and models, the influence (i.e., the absolute value of the regression coefficient) of using \textit{names} and the \textit{interview} format consistently exceeds that of the prompt phrasing, indicating that our main findings are not driven by incidental wording. We further note, that the prompt phrasing shows the least effect on Llama-3.2-70B (no significant coefficients across all measures), while OLMo2-7B is most sensitive to prompt phrasing (significant coefficients for 4 out of 5 measures).

\newcolumntype{d}[1]{S[table-format=#1]}
\begin{table*}[thb]
    \centering
    \small
    \renewcommand\arraystretch{1.1}
    \begin{tabular}{l d{-2.3} d{-2.3} d{-2.3} d{-2.3} d{-2.3}}
    \toprule
     & \multicolumn{1}{c}{Llama-3.3-70B} & \multicolumn{1}{c}{Llama-3.1-8B} & \multicolumn{1}{c}{OLMo-2-32B} & \multicolumn{1}{c}{OLMo-2-7B} & \multicolumn{1}{c}{Gemma-3-27b} \\
    \midrule
    \textcolor{okabe_orange}{Name} & -14.222*** & -6.188*** & -5.639*** & -4.461*** & -18.915*** \\
    \textcolor{okabe_orange}{Structured} & -6.306*** & -2.111*** & -4.300*** & -2.922*** & -5.089* \\
    \textcolor{okabe_red}{Interview} & -11.622*** & -6.099*** & -3.728*** & -3.500*** & -15.264*** \\
    \textcolor{okabe_red}{Third Person} & -1.472 & 1.467** & -0.061 & -1.017* & 2.683 \\
    {Female} & -0.900 & 0.367 & -0.089 & 0.450 & 0.374 \\
    {Nonbinary} & 0.272 & 0.901 & 0.717 & 2.550*** & 0.043 \\
    {Asian} & -0.620 & 0.304 & 0.111 & -1.074 & -2.119 \\
    {Black} & 1.694 & 1.369 & 0.935 & 0.324 & 1.245 \\
    {Hispanic} & 6.250*** & 3.027*** & 6.917*** & 1.102 & 14.220*** \\
    {Middle-Eastern} & 4.056** & 3.101*** & 2.352* & 1.148 & 4.578 \\
    \midrule
    {Self-Description} & 10.333*** & 5.041*** & 4.604*** & 4.448*** & 16.398*** \\
    Prompt Phrasing v2 & 1.089 & 0.441 & 0.974 & 1.478*** & 1.840 \\
    \midrule
    Intercept & 15.548*** & 5.037*** & 2.513* & 2.531*** & 14.448*** \\
    \bottomrule
    \end{tabular}
    \caption{\textbf{Regression on the marked word count.} We conduct OLS regression analyses per LLM using the number of marked words~($\downarrow$) as a dependent variable and report the regression coefficients. The independent variables include: \textcolor{okabe_orange}{demographic priming} (reference: explicit), \textcolor{okabe_red}{role adoption} (reference: direct), {gender} (reference: male), {race} (reference: White), {task} (reference: Bio), and prompt phrasing (reference: v1). *~$p < 0.05$, **~$p < 0.01$, ***~$p < 0.001$.} \label{tab:reg_mw}
\end{table*}

\begin{table*}[thb]
    \centering
    \small
    \renewcommand\arraystretch{1.1}
    \begin{tabular}{l d{-1.3} d{-1.3} d{-1.3} d{-1.3} d{-1.3}}
    \toprule
     & \multicolumn{1}{c}{Llama-3.3-70B} & \multicolumn{1}{c}{Llama-3.1-8B} & \multicolumn{1}{c}{OLMo-2-32B} & \multicolumn{1}{c}{OLMo-2-7B} & \multicolumn{1}{c}{Gemma-3-27b} \\
    \midrule
    \textcolor{okabe_orange}{Name} & 0.068*** & 0.021*** & 0.034*** & 0.036*** & 0.068*** \\
    \textcolor{okabe_orange}{Structured} & -0.006 & -0.009 & 0.002 & 0.007* & -0.001 \\
    \textcolor{okabe_red}{Interview} & 0.009 & 0.041*** & 0.030*** & 0.018*** & 0.006 \\
    \textcolor{okabe_red}{Third Person} & 0.004 & -0.009* & -0.006* & -0.011** & -0.002 \\
    {Female} & -0.006 & -0.020*** & -0.012*** & -0.011** & -0.015** \\
    {Nonbinary} & -0.006 & -0.047*** & -0.023*** & -0.036*** & -0.017** \\
    {Asian} & 0.004 & -0.000 & -0.011*** & -0.023*** & 0.006 \\
    {Black} & 0.004 & -0.005 & -0.017*** & -0.020*** & 0.006 \\
    {Hispanic} & -0.026** & -0.027*** & -0.025*** & -0.018*** & -0.016* \\
    {Middle-Eastern} & -0.004 & -0.016** & -0.024*** & -0.032*** & -0.013* \\
    \midrule
    {Self-Description} & -0.057*** & -0.101*** & -0.128*** & -0.121*** & -0.052*** \\
    Prompt Phrasing v2 & 0.006 & 0.010** & 0.003 & 0.007* & 0.010* \\
    \midrule
    Intercept & 0.300*** & 0.459*** & 0.486*** & 0.506*** & 0.354*** \\
    \bottomrule
    \end{tabular}
    \caption{\textbf{Regression on semantic diversity.} We conduct OLS regression analyses per LLM using semantic diversity~($\uparrow$) as a dependent variable and report the regression coefficients. The independent variables include: \textcolor{okabe_orange}{demographic priming} (reference: explicit), \textcolor{okabe_red}{role adoption} (reference: direct), {gender} (reference: male), {race} (reference: White), {task} (reference: Bio), and prompt phrasing (reference: v1). *~$p < 0.05$, **~$p < 0.01$, ***~$p < 0.001$.} \label{tab:reg_sd}
\end{table*}

\begin{table*}[thb]
    \centering
    \small
    \renewcommand\arraystretch{1.1}
    \begin{tabular}{l d{-1.3} d{-1.3} d{-1.3} d{-1.3} d{-1.3}}
    \toprule
     & \multicolumn{1}{c}{Llama-3.3-70B} & \multicolumn{1}{c}{Llama-3.1-8B} & \multicolumn{1}{c}{OLMo-2-32B} & \multicolumn{1}{c}{OLMo-2-7B} & \multicolumn{1}{c}{Gemma-3-27b} \\
    \midrule
    \textcolor{okabe_orange}{Name} & -0.048*** & -0.027*** & -0.082*** & -0.035*** & -0.040*** \\
    \textcolor{okabe_orange}{Structured} & -0.040*** & -0.018*** & -0.081*** & -0.017* & -0.024* \\
    \textcolor{okabe_red}{Interview} & -0.013 & 0.000 & -0.058*** & -0.033*** & -0.027** \\
    \textcolor{okabe_red}{Third Person} & 0.023* & -0.001 & -0.015 & -0.009 & 0.004 \\
    {Female} & -0.026* & -0.003 & -0.006 & 0.003 & -0.007 \\
    {Nonbinary} & -0.020 & -0.012* & -0.015 & -0.013 & -0.011 \\
    {Asian} & -0.001 & 0.001 & 0.003 & -0.000 & 0.001 \\
    {Black} & -0.000 & -0.000 & 0.006 & 0.002 & -0.000 \\
    {Hispanic} & 0.097*** & 0.053*** & 0.144*** & 0.100*** & 0.094*** \\
    {Middle-Eastern} & -0.001 & -0.002 & 0.021 & -0.001 & 0.001 \\
    \midrule
    {Self-Description} & -0.002 & -0.012** & 0.030** & -0.035*** & 0.024** \\
    Prompt Phrasing v2 & 0.011 & 0.004 & 0.023* & 0.012 & -0.000 \\
    \midrule
    Intercept & 0.038* & 0.028*** & 0.061** & 0.053*** & 0.023 \\
    \bottomrule
    \end{tabular}
    \caption{\textbf{Regression on the share of non-English responses.} We conduct OLS regression analyses per LLM, using the share of non-English responses~($\downarrow$) as a dependent variable and report the regression coefficients. The independent variables include: \textcolor{okabe_orange}{demographic priming} (reference: explicit), \textcolor{okabe_red}{role adoption} (reference: direct), {gender} (reference: male), {race} (reference: White), {task} (reference: Bio), and prompt phrasing (reference: v1). *~$p < 0.05$, **~$p < 0.01$, ***~$p < 0.001$.} \label{tab:reg_lang}
\end{table*}

\begin{table*}[thb]
    \centering
    \small
    \renewcommand\arraystretch{1.1}
    \begin{tabular}{l d{-1.3} d{-1.3} d{-1.3} d{-1.3} d{-1.3}}
    \toprule
     & \multicolumn{1}{c}{Llama-3.3-70B} & \multicolumn{1}{c}{Llama-3.1-8B} & \multicolumn{1}{c}{OLMo-2-32B} & \multicolumn{1}{c}{OLMo-2-7B} & \multicolumn{1}{c}{Gemma-3-27b} \\
    \midrule
    \textcolor{okabe_orange}{Name} & -0.040*** & -0.036*** & -0.017*** & -0.013*** & -0.042*** \\
    \textcolor{okabe_orange}{Structured} & -0.001 & -0.007*** & -0.011*** & -0.009*** & -0.002 \\
    \textcolor{okabe_red}{Interview} & -0.017*** & -0.032*** & -0.024*** & -0.019*** & -0.022*** \\
    \textcolor{okabe_red}{Third Person} & -0.001 & 0.002 & -0.001 & -0.001 & -0.001 \\
    {Female} & -0.003* & -0.004* & -0.004* & -0.003 & -0.005* \\
    {Nonbinary} & -0.008*** & -0.010*** & -0.003 & -0.004* & -0.010*** \\
    {Asian} & -0.009*** & -0.010*** & -0.017*** & -0.017*** & -0.008** \\
    {Black} & -0.005*** & -0.012*** & -0.008*** & -0.010*** & -0.010*** \\
    {Hispanic} & -0.001 & -0.006* & -0.003 & -0.011*** & -0.005 \\
    {Middle-Eastern} & -0.007*** & -0.012*** & -0.015*** & -0.011*** & -0.007** \\
    \midrule
    {Self-Description} & 0.001 & 0.011*** & 0.023*** & 0.015*** & 0.002 \\
    Prompt Phrasing v2 & 0.002 & -0.003 & 0.006*** & 0.003* & 0.003* \\
    \midrule
    Intercept & 1.005*** & 0.982*** & 0.942*** & 0.945*** & 1.005*** \\
    \bottomrule
    \end{tabular}
    \caption{\textbf{Regression on accuracy.} We conduct OLS regression analyses per LLM, using classification accuracy~($\downarrow$) as a dependent variable and report the regression coefficients. The independent variables include: \textcolor{okabe_orange}{demographic priming} (reference: explicit), \textcolor{okabe_red}{role adoption} (reference: direct), {gender} (reference: male), {race} (reference: White), {task} (reference: Bio), and prompt phrasing (reference: v1). *~$p < 0.05$, **~$p < 0.01$, ***~$p < 0.001$.} \label{tab:reg_acc}
\end{table*}

\begin{table*}[thb]
    \centering
    \small
    \renewcommand\arraystretch{1.1}
    \begin{tabular}{l d{-1.3} d{-1.3} d{-1.3} d{-1.3} d{-1.3}}
    \toprule
     & \multicolumn{1}{c}{\textcolor{gray}{Llama-3.3-70B}} & \multicolumn{1}{c}{Llama-3.1-8B} & \multicolumn{1}{c}{OLMo-2-32B} & \multicolumn{1}{c}{OLMo-2-7B} & \multicolumn{1}{c}{\textcolor{gray}{Gemma-3-27b}} \\
    \midrule
    \textcolor{okabe_orange}{Name} & 0.000 & 0.005*** & -0.010*** & 0.003 & 0.001 \\
    \textcolor{okabe_orange}{Structured} & 0.001 & -0.004*** & -0.002 & 0.001 & -0.001 \\
    \textcolor{okabe_red}{Interview} & -0.002 & -0.019*** & -0.013*** & -0.009*** & -0.006*** \\
    \textcolor{okabe_red}{Third Person} & -0.001 & -0.005*** & 0.001 & 0.007*** & -0.001 \\
    {Female} & 0.001 & 0.004*** & 0.002 & 0.001 & 0.000 \\
    {Asian} & -0.031*** & -0.020*** & -0.017*** & -0.016*** & -0.029*** \\
    {Black} & -0.041*** & -0.035*** & -0.038*** & -0.030*** & -0.040*** \\
    {Hispanic} & -0.038*** & -0.026*** & -0.028*** & -0.025*** & -0.037*** \\
    \midrule
    Prompt Phrasing v2 & 0.000 & 0.001 & -0.006*** & -0.003** & -0.001 \\
    \midrule
    Intercept & 0.273*** & 0.196*** & 0.209*** & 0.151*** & 0.276*** \\
    \bottomrule
    \end{tabular}
    \caption{\textbf{Regression on opinion distance.} We conduct OLS regression analyses per LLM, using opinion distance (as measured by Wasserstein distance)~($\downarrow$) as a dependent variable and report the regression coefficients. The independent variables include: \textcolor{okabe_orange}{demographic priming} (reference: explicit), \textcolor{okabe_red}{role adoption} (reference: direct), {gender} (reference: male), {race} (reference: White), and prompt phrasing (reference: v1). We note that the opinion distance of \textcolor{gray}{Llama-3.3-70B} and \textcolor{gray}{Gemma-3-27b} was worse than that of a random baseline. *~$p < 0.05$, **~$p < 0.01$, ***~$p < 0.001$.} \label{tab:reg_dist}
\end{table*}

\subsection{Validation of Log Probabilities for OpinionsQA} \label{app:val_logprobs}

To evaluate whether LLM log probabilities correspond to the model’s actual answer selection behavior, we compare them to answer distributions obtained from multiple generations, where the model selects an option in free-form text. Using OLMo-7B, the best-performing model on OpinionsQA, we prompt each question 100 times with different random seeds and record the answer option chosen in each run. This produces an empirical distribution of answer options, which we compare to the distribution inferred from the model’s log probabilities.
We observe a very low Wasserstein distance (0.034 $\pm$ 0.025) between the log probability distribution and the aggregated answer distribution across 100 runs. This indicates that, for our use case, log probabilities serve as a reliable proxy for the model's actual response behavior

\end{document}